\documentclass[runningheads]{llncs}

\usepackage{eccv}

\usepackage{eccvabbrv}

\usepackage{graphicx}
\usepackage{booktabs}
\usepackage{multirow}

\usepackage{subcaption}
\usepackage[table]{xcolor}
\usepackage{wrapfig}
\usepackage[utf8]{inputenc} 
\usepackage[T1]{fontenc}    
\usepackage{url}            
\usepackage{booktabs}       
\usepackage{amsfonts}       
\usepackage{nicefrac}       
\usepackage{microtype}      
\usepackage{booktabs} 
\usepackage{tabularx}
\usepackage{colortbl}
\usepackage{tablefootnote} 
\usepackage{threeparttable}

\usepackage[accsupp]{axessibility}  
\usepackage{listings}

\usepackage{tikz}
\usepackage{pgfplots}
\pgfplotsset{compat=1.18}
\usepgfplotslibrary{groupplots}

\usepackage[pagebackref,breaklinks,colorlinks,citecolor=eccvblue]{hyperref}
\usepackage{orcidlink}

\makeatletter

\NewDocumentCommand{\linkedfootnotemark}{O{}}{%
    \footnotemark%
    \hypertarget{fnmark:\thefootnote}{}%
    \if\relax\detokenize{#1}\relax\else
        \label{fnmark:#1}%
    \fi
}

\NewDocumentCommand{\linkedfootnotetext}{O{} m}{%
    \footnotetext{%
        \hypertarget{fntext:\thefootnote}{}%
        #2%
    }%
}

\NewDocumentCommand{\linkedfootnote}{O{} m}{%
    \linkedfootnotemark[#1]%
    \linkedfootnotetext[#1]{#2}%
}

\makeatother

\renewcommand{\paragraph}[1]{\noindent \textit{\textbf{#1}}\;}

\definecolor{cityblue}{RGB}{128, 159, 225}

\definecolor{citypink}{RGB}{227, 108, 194}
\definecolor{w_blue}{RGB}{237, 241, 253}
\definecolor{ours}{RGB}{255, 204, 213}
\definecolor{ignore}{gray}{0.85}
\usepackage{array,tabularx}
\newcolumntype{C}{>{\centering\arraybackslash}X}
\definecolor{colorbest}{RGB}{252,187,161}
\definecolor{colorsecond}{RGB}{254,224,210}
\definecolor{colorthird}{RGB}{255,245,240}
\newcommand{\first}[0]{\cellcolor{colorbest}}
\newcommand{\second}[0]{\cellcolor{colorsecond}}
\newcommand{\third}[0]{\cellcolor{colorthird}}
\newcommand{\ignore}[0]{\cellcolor{ignore}}
\DeclareRobustCommand{\legendsquare}[1]{%
  \textcolor{#1}{\rule{2ex}{2ex}}%
}

\newcommand{\ours}{AnyPhoto\xspace}

\begin{document}

\title{AnyPhoto: Multi-Person Identity Preserving Image Generation with ID Adaptive Modulation on Location Canvas} 
\titlerunning{AnyPhoto}
\author{Longhui Yuan}
\authorrunning{L.~Yuan et~al.}
\institute{}

\maketitle
\begin{abstract}
Multi-person identity-preserving generation requires binding multiple reference faces to specified locations under a text prompt.
Strong identity/layout conditions often trigger copy--paste shortcuts and weaken prompt-driven controllability.
We present \ours, a diffusion-transformer finetuning framework with (i) a RoPE-aligned {location canvas} plus location-aligned token pruning for spatial grounding, (ii) AdaLN-style {identity-adaptive modulation} from face-recognition embeddings for persistent identity injection, and (iii) identity-isolated attention to prevent cross-identity interference.
Training combines conditional flow matching with an embedding-space face similarity loss, together with reference-face replacement and location-canvas degradations to discourage shortcuts.
On MultiID-Bench, \ours improves identity similarity while reducing copy--paste tendency, with gains increasing as the number of identities grows. 
\ours also supports prompt-driven stylization with accurate placement, showing great potential application value.
\end{abstract}

\section{Introduction}
In recent years, image generation has witnessed significant progress~\cite{goodfellow2014generative,Ho_Jain_Abbeel_2020,peebles2023scalable,flux2024,esser2024scaling}.
The subfield, identity-preserving image generation, has considerable application value in areas such as social content, film production, and virtual humans, and has consequently attracted considerable research attention~\cite{hyung2024magicapture,papantoniou2024arc2face,valevski2023face0,wang2025stableidentity,wang2024high,wu2024infinite,xiao2025fastcomposer,ye2023ip,gal2022image,guo2024pulid,he2025uniportrait,kim2024instantfamily,zhang2025id,chen2025xverse,wang2024instantid,jiang2025infiniteyou,he2025anystory,mou2025dreamo,xu2025withanyone,qian2025layercomposer,xu2025contextgen,ruiz2023dreambooth,li2024photomaker,yan2023facestudio,zhou2024storymaker}.
ID-preserving image generation aims to synthesize images that follow a text prompt while faithfully rendering one or multiple reference identities. In real applications, identity control is often coupled with explicit spatial constraints: each identity should appear at a designated region in the generated image.

This setting is challenging. First, identity is intrinsically local (mainly carried by facial details), whereas text describes global semantics, requiring the model to balance ROI-specific (Region of Interest) control with global coherence. Second, multi-identity conditioning is prone to cross-identity interference, \ie, different identities may leak into each other without explicit structural constraints. Moreover, when reference faces are constructed from the target image during finetuning, the model may exploit a copy-paste shortcut by directly replicating the reference patch, resulting in visually rigid artifacts.

Existing solutions typically inject spatial control either by explicitly encoding locations (e.g., boxes/masks) as additional conditions~\cite{wang2024instancediffusion,li2025create,wu2025ifadapter}, or by restricting condition-to-image interactions with ROI-masked attention~\cite{zhang2025eligen,wang2024ms,zhou2024storymaker,xiao2025fastcomposer,kim2024instantfamily,he2025uniportrait,he2025anystory}. While effective in certain cases, these designs may increase the learning burden or weaken global-context interactions. Recent findings further suggest that Rotary Position Embedding (RoPE) can provide an implicit mechanism for spatial grounding: when condition tokens share aligned RoPE with image tokens, the model can naturally align conditions to the corresponding regions~\cite{tan2025ominicontrol,tan2025ominicontrol2,qian2025layercomposer,xu2025contextgen}. 

\begin{figure}[t]
    \centering
    \includegraphics[width=0.8\textwidth]{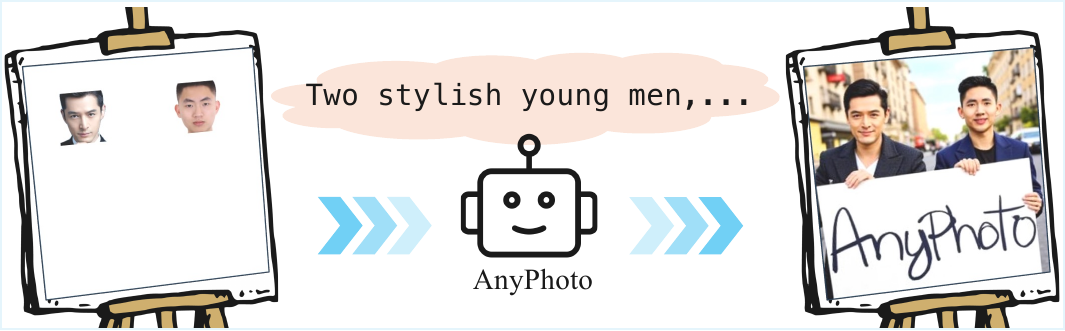}
    \vspace{-3mm}
    \caption{AnyPhoto is capable of generating identity-preserving, location-controlled, high-quality images conditioned on reference faces and a text prompt.}
    \label{fig:intro}
    \vspace{-8mm}
\end{figure}

However, spatial grounding alone does not guarantee identity fidelity. A prevalent approach for identity conditioning is to incorporate an identity embedding as additional input tokens or integrate it directly into the attention mechanism~\cite{xu2025withanyone,valevski2023face0,yan2023facestudio,wang2024instantid,zhou2024storymaker,wu2024infinite,he2025uniportrait}. In practice, this forces the model to learn a nontrivial mapping from a face-recognition embedding space (unseen during pretraining) into the latent generation space, which increases adaptation difficulty, for example, InfU~\cite{jiang2025infiniteyou} requires the introduction of ControlNet-like auxiliary networks along with extensive training data and complex training procedures. Meanwhile, prior work has shown that AdaLN-style modulation provides a more effective conditioning interface in diffusion transformers, significantly amplifying the influence of conditioning signals by controlling feature normalization and residual gating across layers~\cite{chen2025xverse,garibi2025tokenverse}. This motivates us to inject identity not merely as additional tokens, but into the modulation pathway that repeatedly governs the denoising computation, so that the model is encouraged to emphasize identity-invariant cues throughout the network.

Building upon the above observations, we introduce \ours for multi-person identity-preserving image generation with explicit spatial control, as shown in~\cref{fig:intro}.
\ours converts each reference face into {location canvas}
to provide spatially grounded identity evidence, and injects the corresponding identity embedding through {modulation} to strengthen identity control over the designated region.
Specifically, \ours consists of three components:
(1) {Location-Aligned Token Pruning} converts each location canvas into compact ROI tokens with RoPE-aligned coordinates, enabling precise location controllability.
(2) {Identity-Adaptive Modulation} uses a face recognition embedding to modulate AdaLN-style conditioning for the associated identity tokens, enforcing consistent identity emphasis across transformer blocks.
(3) {Identity-Isolated Attention} structures attention connectivity so that global tokens coordinate all conditions while each identity branch remains disentangled, supporting stable multi-person composition.
We finetune \ours with a combined objective of conditional flow matching and an embedding-space face similarity loss, which jointly optimizes image realism and identity fidelity, and mitigate copy-paste collapse via same-ID reference replacement and location-canvas degradations.

{The contributions of this work can be summarized as follows:}
\begin{itemize}
    \item We propose \ours, a finetuning framework generates {identity-preserving}, {location-controlled}, and {high-quality} images for multi-person scenarios.
    \item We introduce Location-Aligned Token Pruning for RoPE-aligned ROI representation, Identity-Adaptive Modulation for stronger identity-specific control, and Identity-Isolated Attention for disentangled multi-identity interaction.
    \item Experiments demonstrate that \ours achieves higher face similarity and more accurate spatial placement in multi-person generation, while producing more natural results with fewer copy-paste artifacts.
\end{itemize}
\section{Preliminaries}
\subsection{Flow Matching}
Flow matching trains generative models by learning a time-dependent velocity field transporting a simple noise distribution $q_1$ to the data distribution $q_0$, and it underlies modern systems such as Stable Diffusion 3~\cite{esser2024scaling} and FLUX~\cite{flux2024}.
Concretely, flow-based generative modeling defines an ordinary differential equation (ODE)
\begin{equation}
    \frac{d}{dt}x(t)=\mu(x(t),t),
\end{equation}
where $x(1)\sim q_1$ and integrating the dynamics backward to $t=0$ yields $x(0)\sim q_0$.
Flow matching aims to learn a parameterized vector field $\mu_\theta(x,t)$ whose induced probability path matches a target path.
Conditional Flow Matching (CFM)~\cite{lipman2022flow} solves the problem by constructing a conditional probability path $p_t(x_t\mid x_0)$, which induces the marginal path
\begin{equation}
    p_t(x_t)=\int p_t(x_t\mid x_0) q_0(x_0)\,dx_0,
\end{equation}
and satisfies the boundary conditions $p_0=q_0$ and $p_1=q_1$.

Let $\mu^*(x_t,t\mid x_0)$ denote a conditional vector field that generates the conditional path $p_t(x_t\mid x_0)$, and define the corresponding marginal vector field by
\begin{equation}
    \mu^*(x_t,t)=\int \mu^*(x_t,t\mid x_0)\,\frac{p_t(x_t\mid x_0)q_0(x_0)}{p_t(x_t)}\,dx_0.
\end{equation}
Lipman~\etal~\cite{lipman2022flow} show that if $\mu^*(x_t,t\mid x_0)$ generates $p_t(x_t\mid x_0)$, then $\mu^*(x_t,t)$ generates the marginal path $p_t(x_t)$.
Accordingly, flow matching minimizes the mean squared error between the learned field and the target field:
\begin{align}
    \mathcal{L}_{FM}(\theta) &= \mathbb{E}_{t,\,x_t\sim p_t}\Big[\big\|\mu_\theta(x_t,t)-\mu^*(x_t,t)\big\|^2\Big], \\
    \mathcal{L}_{CFM}(\theta) &= \mathbb{E}_{t,\,x_0\sim q_0,\,x_t\sim p_t(\cdot\mid x_0)}\Big[\big\|\mu_\theta(x_t,t)-\mu^*(x_t,t\mid x_0)\big\|^2\Big].
\end{align}
Moreover, $\mathcal{L}_{FM}(\theta)$ and $\mathcal{L}_{CFM}(\theta)$ share the same gradient with respect to $\theta$, so one can train using the conditional objective without explicitly computing the intractable marginal target field.

Rectified Flow (RF)~\cite{liu2022flow} can be viewed as a special case of CFM obtained by choosing a deterministic linear probability path.
Specifically, sample $x_0\sim q_0$ and $x_1\sim q_1$, and define for $t\in[0,1]$ the interpolation
\begin{equation}
    x_t=(1-t)x_0+tx_1.
\end{equation}
In this case, the conditional dynamics are deterministic and the corresponding conditional vector field is simply
\begin{equation}
    \mu^*(x_t,t\mid x_0,x_1)=\frac{d}{dt}x_t=x_1-x_0.
\end{equation}
Substituting this choice into the conditional flow matching objective yields the rectified flow training loss
\begin{equation}
    \mathcal{L}_{RF}(\theta)
    =\mathbb{E}_{t\sim\mathcal{U}[0,1],\,x_0\sim q_0,\,x_1\sim q_1}
    \left[\left\|\mu_\theta\left(x_t,t\right)-(x_1-x_0)\right\|^2\right].
\end{equation}
After training, samples are generated by solving the ODE from $t=1$ to $t=0$ with the learned vector field $\mu_\theta$ and initial condition $x(1)\sim q_1$.

\subsection{Modulation in DiTs}
Modern diffusion transformers commonly inject conditions through {AdaLN-style modulation} with gated residual updates, a design that traces back to StyleGAN~\cite{Karras_Laine_Aila_2019}. Importantly, this design explicitly {strengthens the influence of conditioning} by modulating normalized activations at every block, so that the prompt reliably steers the denoising dynamics across layers and timesteps.
Stable Diffusion 3~\cite{esser2024scaling} and FLUX~\cite{flux2024} adopt this mechanism: timestep and text embeddings are fused and mapped to global conditioning vector
\begin{equation}
    y = \text{MLP}(t, f_{prompt}).
\end{equation}
A following MLP produces per-block (or per-sublayer) modulation parameters $[\alpha,\beta,\gamma]$. For an activation $x$, AdaLN applies a feature-wise affine transformation to the LayerNorm output,
\begin{equation}
    \tilde{x} \;=\; (1+\alpha)\odot \mathrm{LN}(x) + \beta,
\end{equation}%
which is then processed by the attention/MLP sublayer $F(\cdot)$ and injected via a gated residual connection,
\begin{equation}
    x \;\leftarrow\; x + \gamma \odot F(\tilde{x}).
\end{equation}%
Here, $(\alpha,\beta)$ provide StyleGAN-like feature scaling and shifting, while $\gamma$ controls the update strength under the current timestep and prompt.

\section{\ours}

\begin{figure}[t]
    \centering
    \includegraphics[width=0.75\textwidth]{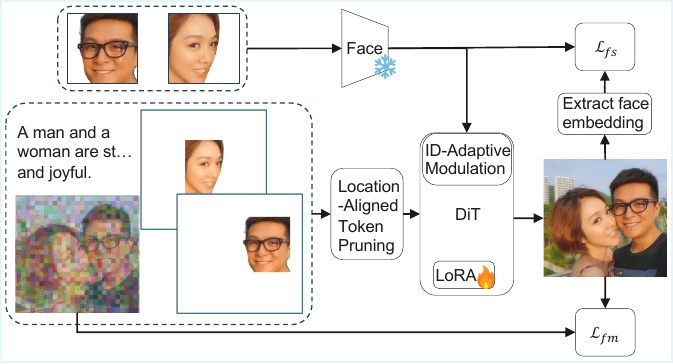}
    \vspace{-4mm}
    \caption{
        AnyPhoto. Location-Aligned Token Pruning constructs the input from the text prompt, noised image, and location reference. Reference face embeddings modulate the aligned tokens. Training uses conditional flow matching and face similarity losses.}
    \label{fig:framework}
    \vspace{-7mm}
\end{figure}

\label{sec:methods}
\subsection{Problem Definition}
We study identity-preserving image generation with explicit spatial constraints.
Given a text prompt $s$ and a set of $n$ reference pairs
$
\mathcal{R}=\{(r_i,l_i)\}_{i=1}^{n},
$
where $r_i$ is the $i$-th reference face image (norm-cropped) and $l_i$ specifies its target location (e.g., face keypoints) in the generated image, our goal is to synthesize an image $x$ that (i) is semantically consistent with $s$, and (ii) renders each identity $r_i$ at its designated location $l_i$.
Formally, we aim to learn a conditional generative model parameterized by $\theta$ that approximates the target distribution
\begin{equation}
p_\theta(x \mid s, \mathcal{R}),
\end{equation}
so that samples $x \sim p_\theta(\cdot \mid s, \mathcal{R})$ satisfy both the textual description and the location-aligned multi-identity constraints.
In this work, we propose the \ours method to achieve this goal, and an overview is shown in~\cref{fig:framework}. 

\subsection{Location-Aligned Token Pruning}
In prior work, explicit positional control is typically achieved in two ways. One is to encode spatial cues (e.g., coordinates, or boxes) as additional inputs\cite{wang2024instancediffusion,li2025create,wu2025ifadapter}, which is simple but increases the learning burden because the model must learn both the target semantics and how to interpret these spatial representations. The other is ROI-masked attention\cite{zhang2025eligen,wang2024ms,zhou2024storymaker,xiao2025fastcomposer,kim2024instantfamily,he2025uniportrait,he2025anystory}, which localizes control by restricting attention between condition tokens and ROI tokens, but may weaken global-context interactions and lead to visually inconsistent or fragmented results.

Recent work such as OmniControl~\cite{tan2025ominicontrol, tan2025ominicontrol2} reveals the importance of Rotary Position Embedding (RoPE) for condition tokens. When a condition map (e.g., depth, canny) matches the target image resolution, assigning {aligned} RoPE to condition tokens yields strong spatial control. This suggests that RoPE provides an implicit spatial alignment mechanism: sharing same rotational basis makes condition tokens comparable to image tokens in positional phase, enabling region-level alignment without explicit attention masking or extra positional encoding.

To fully exploit this implicit spatial controllability, we propose a {Location-Aligned Token Pruning} module. 
Given a norm-cropped reference face image $r_i$ and its target location $l_i$, we first construct a {location canvas} $L_i$ by resizing $r_i$ according to $l_i$ and pasting it onto an otherwise blank white canvas, which has the same resolution as the target image. 
This operation also yields a binary pasted-region mask $M_i$ indicating the spatial support of the reference face on the canvas. 
We then encode $L_i$ with the VAE encoder to obtain a dense latent feature map (tokens) $F_i$. 
Next, we downsample $M_i$ to the token resolution and remove background (blank) tokens by indexing with the mask,
\begin{equation}
F_i \leftarrow F_i[M_i],
\end{equation}
thereby retaining only tokens that correspond to the pasted face region and discarding tokens from the blank canvas.

In parallel, we construct a 3-D RoPE coordinate for each token as $[1, x, y]$, where the leading dimension indicates the token type (location condition) and $(x,y)$ are the spatial coordinates. 
We prune RoPE coordinates using the same mask to obtain the RoPE assigned to the retained face tokens.

\begin{figure}[t]
    \centering
    \includegraphics[width=\textwidth]{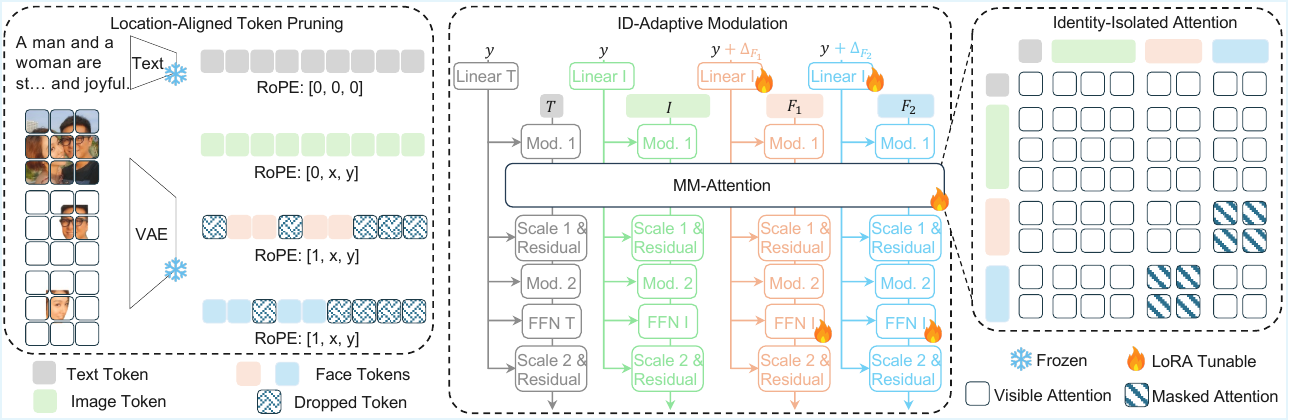}
    \vspace{-7mm}
    \caption{Detailed components illustration of AnyPhoto.}
    \label{fig:method}
    \vspace{-7mm}
\end{figure}

For the remaining inputs, we encode the text prompt using a text encoder to obtain text tokens $T$ and assign them RoPE coordinates $[0,0,0]$. 
We encode the noised image using the VAE to obtain visual tokens $I$ with RoPE coordinates $[0,x,y]$. 
Finally, we concatenate all tokens as
\begin{equation}
X = [T, I, \{F_i\}_{i=1}^{n}],
\end{equation}
and likewise concatenate their corresponding RoPE coordinates, enabling the model to attend over text, image, and location-aligned identity tokens within a unified rotary positional framework.
And since each face is processed individually, the model can learn the mutual occlusion relationships between different subjects.
We graphically illustrate this process on the left side of~\cref{fig:method}.

\subsection{Identity-Adaptive Modulation}
Location-Aligned Token Pruning enables the model to {place} identity evidence at the desired spatial region by providing location-aligned identity tokens $F_i$. 
Yet, preserving identity is not only determined by whether the model can attend to $F_i$, but also by whether the network consistently emphasizes the {identity-invariant cues} contained in $F_i$ (e.g., facial structure and characteristic appearance) across the denoising trajectory. 
In diffusion transformers, such persistent emphasis is naturally implemented through AdaLN-style modulation, where a conditioning vector is repeatedly mapped to per-block scale/shift/gating parameters that control feature statistics and residual strength. 
Motivated by this, we inject identity into the modulation pathway, so that identity-related processing is encouraged at every block rather than being left to emerge implicitly from attention.
Specifically, for each reference face $r_i$, we extract an identity embedding using a pretrained face recognition network (e.g., ArcFace~\cite{deng2019arcface}),
\begin{equation}
e_{r_i} = \mathrm{FaceNet}(r_i).
\end{equation}
We then map $e_{r_i}$ to an additive modulation offset through an MLP,
\begin{equation}
\Delta_{F_i} = \mathrm{MLP}_{\mathrm{id}}(e_{r_i}),
\end{equation}
where $\Delta_{F_i}$ has the same dimensionality as the transformer conditioning vector $y$ (derived from timestep and text). 
During the forward pass, whenever a modulation layer operates on tokens originating from $F_i$, we use an identity-adapted conditioning vector
\begin{equation}
y_i = y + \Delta_{F_i},
\end{equation}
and compute the corresponding modulation parameters from $y_i$,
\begin{equation}
[\alpha,\beta,\gamma] = \mathrm{MLP}_{\mathrm{mod}}(y_i), \qquad \text{for tokens in } F_i.
\end{equation}
Because AdaLN modulation directly controls normalized feature scaling and the effective residual update, adding $\Delta_{F_i}$ biases the network to allocate more capacity to those components of $F_i$ that are predictive of the target identity, while suppressing spurious, identity-irrelevant variations. 
As a result, identity information is reinforced in a layer-by-layer manner, yielding stronger identity consistency without globally overwriting non-identity tokens and without requiring additional mask supervision. For a more intuitive understanding, please refer to the middle part of~\cref{fig:method}.

\subsection{Identity-Isolated Attention}
We introduce an {Identity-Isolated Attention} mask to prevent cross-identity interference. The input sequence contains text tokens $T$, noised image tokens $I$, and $n$ sets of location-aligned identity tokens $\{F_i\}_{i=1}^{n}$. We let the global tokens $(T,I)$ gather information from all sources, while restricting each identity branch $F_i$ to interact only with $(T,I)$ and itself.
Concretely, the attention rule is:
(i) $T$ and $I$ attend to all tokens in $[T, I, F_1,\dots,F_n]$;
(ii) $F_i$ attends to $T$, $I$, and $F_i$, but not to $F_j$ for $j\neq i$.
This forms a star-shaped topology centered at $(T,I)$.
Formally, for $X=[T,I,F_1,\dots,F_n]$ and mask $A\in\{0,1\}^{|X|\times|X|}$,
\begin{equation}
A_{pq} =
\begin{cases}
1, & p\in (T\cup I),\\
1, & p\in F_i \ \text{and}\ q\in (T\cup I\cup F_i),\\
0, & \text{otherwise},
\end{cases}
\end{equation}
where $p$ and $q$ index the query and key/value tokens.
This design preserves global coherence while explicitly isolating different identities (see the right of~\cref{fig:method}).

\subsection{\ours Training}

We finetune our model with a combined objective consisting of a conditional flow matching loss and a face similarity regularization.
Given a training sample $(x, s, \mathcal{R})$, where $x$ is the target image, $s$ is the text prompt, and $\mathcal{R}=\{(r_i,l_i)\}_{i=1}^{n}$ are identity-location pairs, we adopt conditional flow matching to learn the mapping from the newly introduced conditions to the target image distribution.

Let $x_0 \sim p_{\mathrm{data}}$ and $x_1 \sim p_{\mathrm{noise}}$ denote a data sample and a noise sample, respectively.
We consider the linear interpolation path
$x_t = (1-t)x_0 + tx_1$ with $t\sim \mathcal{U}(0,1)$, and train a velocity field $\mu_\theta(\cdot)$ conditioned on $(s,\mathcal{R})$.
The conditional flow matching loss is written as
\begin{equation}
\mathcal{L}_{{fm}} =
\mathbb{E}_{t, x_0,x_1,s,\mathcal{R}}\Big[
\big\|
\mu_\theta(x_t, s, \mathcal{R}, t) - (x_1 - x_0)
\big\|_2^2
\Big].
\end{equation}
Optimizing $\mathcal{L}_{{fm}}$ encourages the model to capture how the additional identity-and-location conditions translate into the target image.

\paragraph{Avoiding copy-paste collapse.}
We observe that if reference faces are always constructed from the same image $x$ being reconstructed, the model easily falls into a copy-paste shortcut, i.e., directly replicating the reference face patches onto the specified locations, leading to visually rigid and unnatural generations.
To mitigate this, for each $(r_i,l_i)\in\mathcal{R}$ we randomly replace $r_i$ with another face $\tilde{r}_i$ sampled from a reference face bank that belongs to the same identity as $r_i$, forming an augmented set
$\tilde{\mathcal{R}}=\{(\tilde{r}_i,l_i)\}_{i=1}^{n}$.
Meanwhile, when constructing the location canvas $L_i$ (used to derive $F_i$), we apply a series of degradations (e.g., blur, color jitter and so on) to the pasted face region, improving robustness and discouraging trivial pixel-level copying.
Ideally, faces of the same identity yield similar ArcFace embedding; thus, through ID-adaptive modulation, the model is encouraged to attend to identity-invariant cues within $F_i$ rather than superficial appearance variations.
It is worth noting that, following the concept of curriculum learning~\cite{bengio2009curriculum}, we gradually and incrementally increase the face replacement probability during the training process to achieve end-to-end training.

\paragraph{Face similarity loss.}
Nevertheless, $\mathcal{L}_{{fm}}$ alone mainly encourages reconstruction and may not sufficiently force the model to preserve identity-discriminative characteristics.
We therefore introduce an additional face similarity loss based on ArcFace embeddings.
Given
$x_t = (1-t)x_0 + tx_1$ and
\begin{equation}
\mu_t = \mu_\theta(x_t, s, \tilde{\mathcal{R}}, t),
\end{equation}
we obtain a one-step estimate of the clean image
\begin{equation}
\hat{x}_0 = x_t - t\,\mu_t.
\end{equation}
For each location $l_i$, we obtain $\hat{r}_i$ by
\begin{equation}
\hat{r}_i = \mathrm{CROP}(\hat{x}_0, l_i).
\end{equation}
We then compute ArcFace embeddings for both the reference face and the predicted face,
$e_{r_i}=\mathrm{FaceNet}(r_i)$ and $e_{\hat{r}_i}=\mathrm{FaceNet}(\hat{r}_i)$, and define the cosine-distance loss as
\begin{equation}
\mathcal{L}_{{fs}} = \frac{1}{n}\sum_{i=1}^{n}\Big(1 - \cos(e_{r_i}, e_{\hat{r}_i})\Big).
\end{equation}

\paragraph{Overall objective.}
The final training objective is
\begin{equation}
\mathcal{L} = \mathcal{L}_{{fm}} + \lambda\,\mathcal{L}_{{fs}},
\end{equation}
where $\lambda$ controls the strength of the identity-preservation regularization.

\section{Experiments}
\label{sec:experiments}

\subsection{Experimental Setup}
\label{sec:exp_setup}
\paragraph{Datasets.} 
\ours is finetuned on MultiID-2M~\cite{xu2025withanyone}, a large-scale dataset of multi-identity images paired with identity reference images. 
During preprocessing, we filter out samples that do not meet the data requirements (e.g., missing/low-quality references) or fail in our processing pipeline, resulting in about 0.7M images used for training. 
We evaluate \ours on MultiID-Bench~\cite{xu2025withanyone}. For each reference identity, we detect the target subject location in the reference image and use it as the spatial control condition for location guidance.

\paragraph{Baselines.}
Compare with representative methods, including,
(1) General customization models, DreamO~\cite{mou2025dreamo}, OmniGen~\cite{xiao2025omnigen}, OminiGen2~\cite{wu2025omnigen2}, Qwen-Image-Edit~\cite{wu2025qwen}, FLUX.1 Kontext~\cite{labs2025flux},UNO~\cite{wu2025less}, USO~\cite{wu2025uso}, UMO~\cite{cheng2025umo}, and native GPT-4o-Image~\cite{hurst2024gpt}. (2) face customization methods, InfU~\cite{jiang2025infiniteyou}, UniPortrait~\cite{he2025uniportrait}, ID-Patch~\cite{zhang2025id}, PuLID~\cite{guo2024pulid}, InstantID~\cite{wang2024instantid}, WithAnyOne~\cite{xu2025withanyone}.

\paragraph{Implementation Details.} We build \ours upon FLUX.1-[dev]~\cite{flux2024}. To accommodate the newly introduced condition tokens, we attach a LoRA~\cite{hu2022lora} module (rank $r{=}128$) to the image branch, and use an MLP to process the ArcFace embedding. We optimize the model with AdamW (lr=$5\times10^{-4}$, betas=$(0.9,0.95)$, weight decay=$10^{-3}$). We further adopt a curriculum for reference-face replacement, where the replacement probability is increased from 0 to 0.5 following the schedule $[0.0, 0.05, 0.1, 0.2, 0.3, 0.4, 0.5]$ at milestone steps $[10\mathrm{k}, 20\mathrm{k}, 30\mathrm{k}, 40\mathrm{k}, 50\mathrm{k}, 60\mathrm{k}]$. The loss balancing weight is set to $\lambda{=}0.1$. 
To enable classifier-free guidance (CFG)~\cite{ho2022classifier} at inference, we drop the identity conditions during training by setting both the location tokens and the ArcFace embeddings of each identity to zero with probability 15\%. 
We train for 200k steps with batch size 1 per GPU on 64 NVIDIA H20 GPUs \begin{wraptable}{r}{0.6\textwidth}
    \centering
    \vspace{-3mm}
        \caption{
        Comparison results on MultiID-Bench. \legendsquare{colorbest}, \legendsquare{colorsecond}, and \legendsquare{colorthird} denote the 1st/2nd/3rd best results. For Copy-Paste/Generation-Quality ranking, we only include cases with $\mathrm{Sim(GT)} > 0.40$/$\mathrm{Sim(Ref)} > 0.50$. \legendsquare{ignore} marks ignored cases in the ranking.
        }
        \scalebox{0.6}{
        \renewcommand{\arraystretch}{0.3}
        \begin{tabular}{lcccccc}
        \toprule
        \toprule
        \multirow{2}[2]{*}{\textbf{Method}} & \multicolumn{3}{c}{\textbf{Identity Metrics}} & \multicolumn{3}{c}{\textbf{Generation Quality}} \\
        \cmidrule(r){2-4} \cmidrule(l){5-7}
         & Sim(GT) $\uparrow$ & Sim(Ref) $\uparrow$ &  CP $\downarrow$ & CLIP-I $\uparrow$ & CLIP-T $\uparrow$ & Aes $\uparrow$ \\
        \midrule
        \multicolumn{7}{c}{1-people subset} \\
        \midrule
        DreamO & 0.454 & 0.694 & 0.303 & \third{0.793} & \third{0.322} & 4.877 \\
        OmniGen & 0.398 & 0.602 & \ignore 0.248 & 0.780 & 0.317 & 5.069 \\
        OmniGen2 & 0.365 & 0.475 & \ignore 0.142 & \ignore 0.787 & \ignore{0.331} & \ignore 4.991 \\
        FLUX.1 Kontext & 0.324 & 0.408 & \ignore 0.099 & \ignore 0.755 & \ignore 0.327 & \ignore{5.319} \\
        Qwen-Image-Edit & 0.324 & 0.409 & \ignore 0.093 & \ignore 0.776 & \ignore 0.316 & \ignore 5.056 \\
        GPT-4o-Image & 0.425 & 0.579 & \third{0.178} & \second{0.794} & 0.311 & \second{5.344} \\
        UNO & 0.304 & 0.428 & \ignore 0.141 & \ignore 0.765 & \ignore 0.314 & \ignore 4.923 \\
        USO & 0.401 & 0.635 & 0.286 & 0.790 & \first{0.329} & 5.077 \\
        UMO & \third{0.458} & \second{0.732} & 0.359 & 0.783 & 0.305 & 4.850 \\
        \hline
        UniPortrait & 0.447 & 0.677 & 0.265 & \third{0.793} & 0.319 & 5.018 \\
        ID-Patch & 0.426 & 0.633 & {0.231} & 0.792 & 0.312 & 4.900 \\
        InfU & 0.439 & 0.630 & 0.233 & 0.772 & \second{0.328} & \first{5.359} \\
        PuLID  & 0.452 & \third{0.705} & 0.315 & 0.779 & 0.305 & 4.839 \\
        InstantID  & \first{0.464} & \first{0.734} & 0.337 & 0.764 & 0.295 & \third{5.255} \\
        WithAnyone & \second{0.460} & 0.578 & \first{0.144} & \first{0.798} & 0.313 & 4.783 \\
        \ours & 0.448 & 0.600 & \second{0.176} & 0.788 & 0.315 & 5.011 \\
        \hline
        \rowcolor{blue!5} GT & 1.000 & 0.521 & -0.999 & N/A & N/A & N/A \\
        \rowcolor{blue!5} Ref & 0.521 & 1.000 & 0.999 & N/A & N/A & N/A \\
        \midrule
        \multicolumn{7}{c}{2-people subset} \\
        \midrule
        DreamO & {0.359} & 0.514 & {\ignore 0.179} & 0.763 & 0.319 & 4.764 \\
        OmniGen & 0.345 & {0.529} & \ignore 0.209 & 0.750 & \first{0.326} & \second{5.152} \\
        OmniGen2 & 0.283 & 0.353 & \ignore 0.081 & \ignore 0.763 & \ignore{0.334} & \ignore 4.547 \\
        GPT-4o-Image & 0.332 & 0.400 & \ignore 0.061 & \ignore{0.774} & \ignore{0.328} & \ignore{5.676} \\
        UNO & 0.223 & 0.274 & \ignore 0.043 & \ignore 0.735 & \ignore 0.325 & \ignore 4.805 \\
        \hline
        UMO& 0.328 & 0.491 & \ignore 0.176 & \ignore 0.743 & \ignore 0.316 & \ignore 4.772 \\
        UniPortrait & \third{0.367} & \first{0.601} & \ignore 0.254 & 0.750 & \second{0.323} & \first{5.187} \\
        ID-Patch & 0.350 & 0.517 & {\ignore 0.183} & \second{0.767} & \first{0.326} & 4.671 \\
        WithAnyone & \first{0.405} & \third{0.551} & \first{0.161} & \first{0.770} & \third{0.321} & 4.883 \\
        \ours & \second{0.401} & \second{0.577} & \second{0.189} & \third{0.764} & \second{0.323} & \third{4.994} \\
        \midrule
        \multicolumn{7}{c}{3-and-4-people subset} \\
        \midrule
        DreamO      & 0.311 & 0.427 & \ignore 0.116 & \ignore 0.709 & \ignore 0.317 & \ignore 4.695 \\
        OmniGen     & {0.345} & {0.529} & \ignore 0.209 & 0.750 & \second{0.326} & \first{5.152} \\
        OmniGen2    & 0.288 & 0.374 & \ignore 0.099 & \ignore 0.734 & \ignore{0.329} & \ignore 4.664 \\
        GPT-4o-Image~\linkedfootnotemark[2] & \ignore \textit{0.445} & \ignore \textit{0.484} & \ignore \textit{0.048} & \ignore {0.815} & \ignore {0.320} & \ignore {5.647} \\
        UNO         & 0.228 & 0.276 & \ignore 0.046 & \ignore 0.717 & \ignore 0.319 & \ignore 4.880 \\
        \hline
        UMO         & 0.318 & 0.465 & \ignore 0.180 & \ignore 0.717 & \ignore 0.309 & \ignore 4.946 \\
        
        UniPortrait & 0.343 & 0.517 & \ignore 0.178 & 0.708 & 0.323 & \second{5.090} \\
        ID-Patch    & \third{0.379} & \third{0.543} & {\ignore 0.195} & \first{0.781} & \first{0.329} & 4.547 \\
        WithAnyone        & \second{0.414} & \second{0.561} & \first{0.171} & \second{0.771} & \third{0.325} & 4.955 \\
        \ours & \first{0.424} & \first{0.587} & \second{0.186} & \third{0.763} & \second{0.326} & \third{4.960} \\
        \bottomrule
        \bottomrule
    \end{tabular}
        }       
    \label{tab:quantitative_comparison}
    \vspace{-17mm}
\end{wraptable}%
\linkedfootnotetext[2]{We follow WithAnyone to exclude GPT in the similarity ranking.}%
 (global batch size 64), which takes about 5 days.
For inference, the CFG scale is set to 1.0 (effectively disabling it) with 28 sampling steps.

\subsection{Main Results}
\label{sec:main_results}
\paragraph{Quantitative Results.}
\cref{tab:quantitative_comparison} compares \ours with baselines on MultiID-Bench, covering identity similarity $\quad$ (Sim(GT/Ref)), copy-paste tendency (CP), and generation quality (CLIP-I/CLIP-T/Aes; see the ranking rules in the caption). 
Notably, the GT/Ref rows indicate a non-negligible gap between the two similarity references, so strong performance requires balancing both similarity scores while suppressing copy-paste behavior.
Overall, \ours achieves the most robust balance across subsets, especially as the number of identities increases. On the challenging 3--4-people subset, \ours ranks \textbf{1st} on both similarity scores, while remaining among the best methods in CP, demonstrating that our identity control scales to crowded scenes without collapsing into reference pasting. 
On the 2-people subset, \ours stays consistently in the \textbf{top-2} for both similarity scores
and maintains low CP, reflecting stable controllability when multiple identities compete for attention. On the 1-people subset, some face-specialized methods attain higher similarity but also show more pronounced copy-paste tendency; in contrast, \ours preserves strong identity similarity while keeping CP among the lowest, indicating more compositional synthesis rather than direct copying. Meanwhile, \ours maintains competitive CLIP-T and aesthetic scores across all subsets, suggesting that improved identity controllability does not compromise prompt alignment or visual realism.

\begin{figure}[t]
    \centering
    \includegraphics[width=\textwidth]{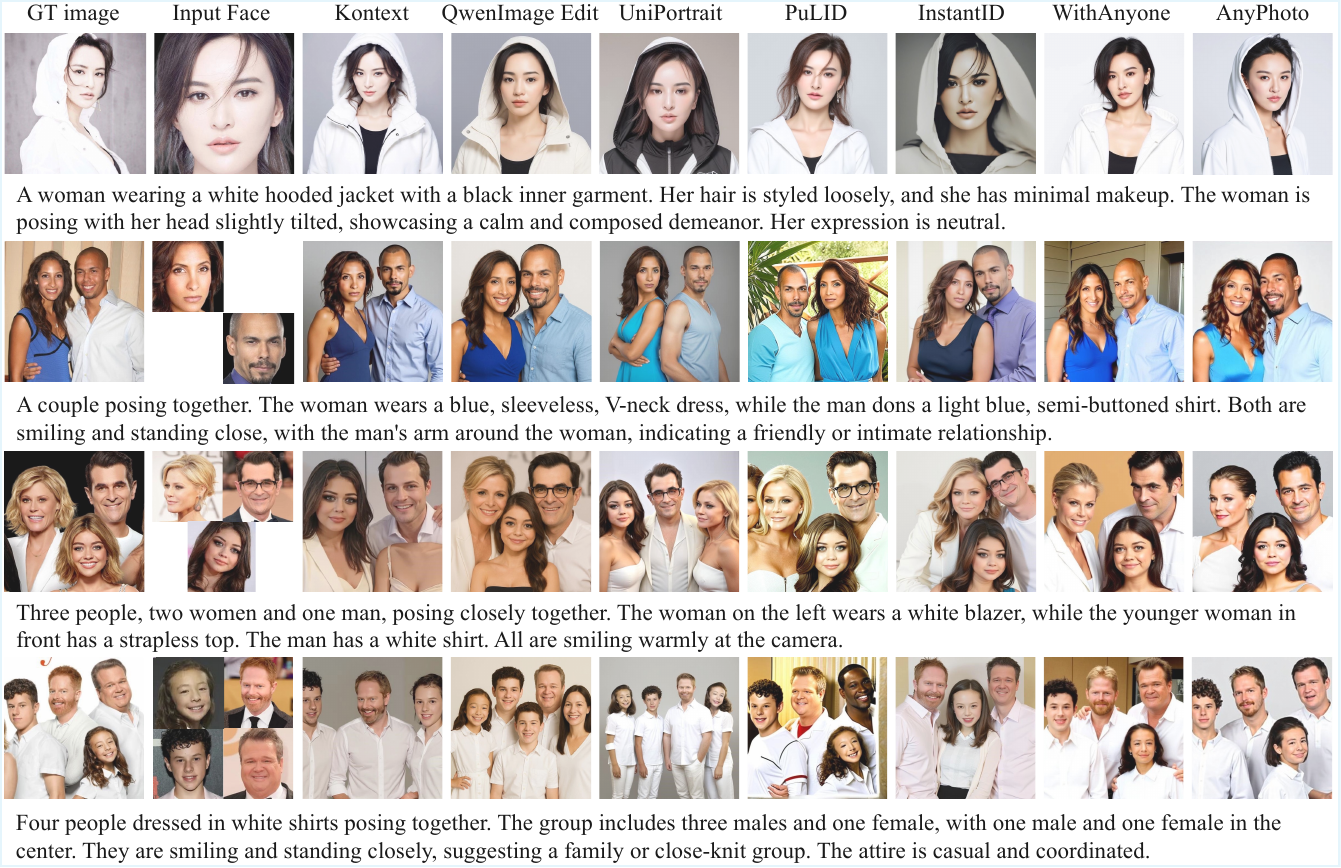}
    \vspace{-7mm}
    \caption{Visual comparisons of \ours with baselines conditioned on 1/2/3/4 persons.}
    \label{fig:compare_baselines}
    \vspace{-7mm}
\end{figure}

\paragraph{Qualitative Comparisons.}
\cref{fig:compare_baselines} compares \ours with representative baselines under 1/2/3/4-identity conditioning. We observe several common failure modes in existing methods, including missing identities, hallucinating extra people, drifting away from the specified locations, and obvious copy-paste artifacts---issues that typically become more severe as the number of identities increases. In contrast, \ours produces consistently good results across these cases, preserving each identity while keeping them in the desired regions with fewer artifacts and more coherent compositions.

\subsection{Ablation Studies}
\label{sec:ablation}
\begin{figure*}[t]
    \centering
    \begin{subfigure}[t]{0.13\textwidth}
        \centering
        \includegraphics[width=\linewidth]{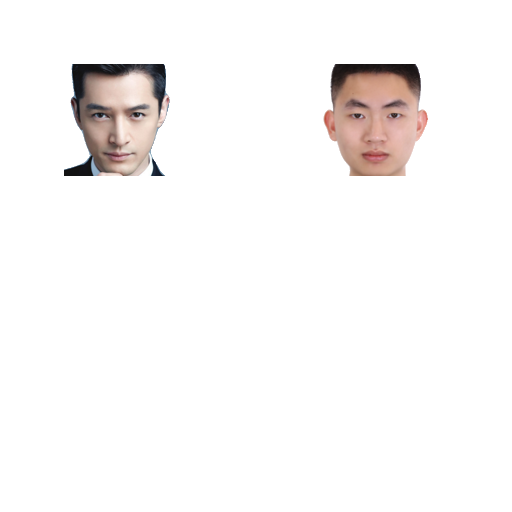}
    \end{subfigure}\hfill
    \begin{subfigure}[t]{0.13\textwidth}
        \centering
        \includegraphics[width=\linewidth]{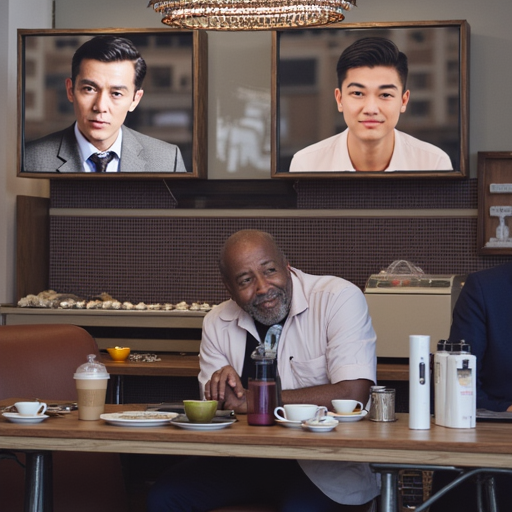}
    \end{subfigure}\hfill
    \begin{subfigure}[t]{0.13\textwidth}
        \centering
        \includegraphics[width=\linewidth]{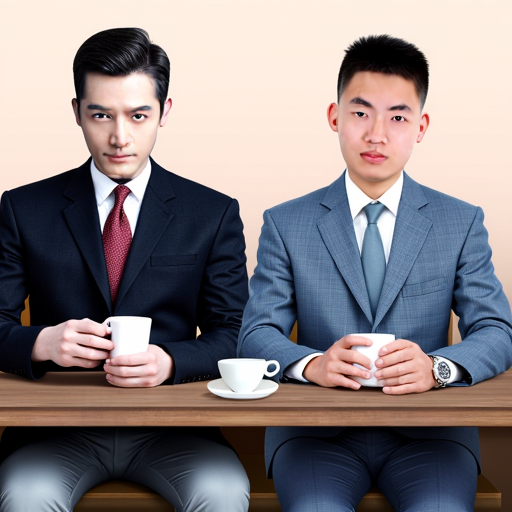}
    \end{subfigure}\hfill
    \begin{subfigure}[t]{0.13\textwidth}
        \centering
        \includegraphics[width=\linewidth]{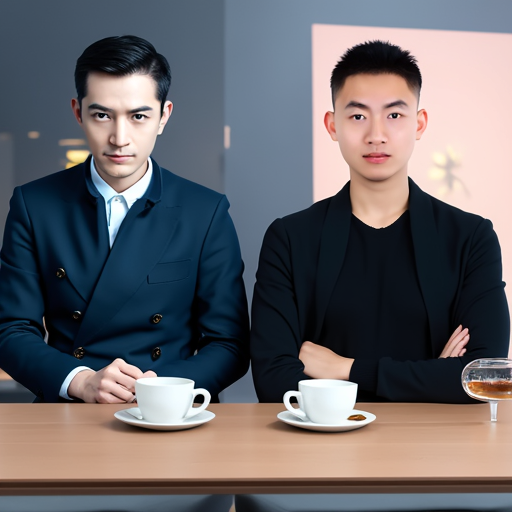}
    \end{subfigure}\hfill
    \begin{subfigure}[t]{0.13\textwidth}
        \centering
        \includegraphics[width=\linewidth]{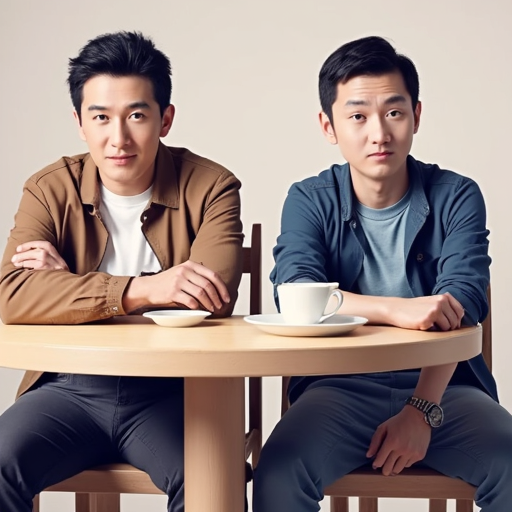}
    \end{subfigure}\hfill
    \begin{subfigure}[t]{0.13\textwidth}
        \centering
        \includegraphics[width=\linewidth]{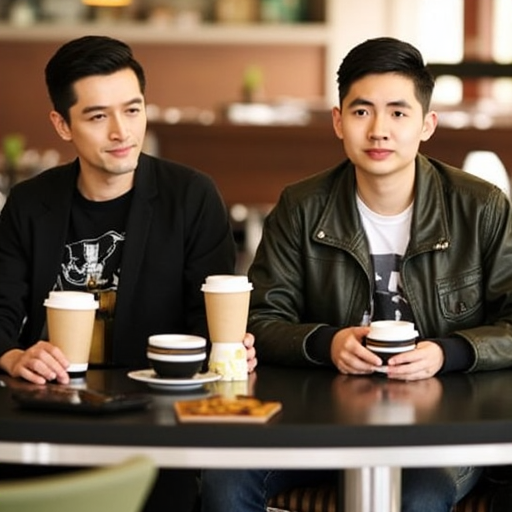}
    \end{subfigure}\hfill
    \begin{subfigure}[t]{0.13\textwidth}
        \centering
        \includegraphics[width=\linewidth]{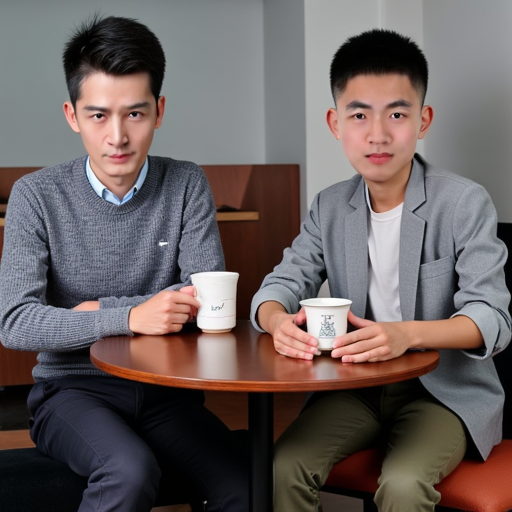}
    \end{subfigure}

    \begin{subfigure}[t]{0.13\textwidth}
        \centering
        \includegraphics[width=\linewidth]{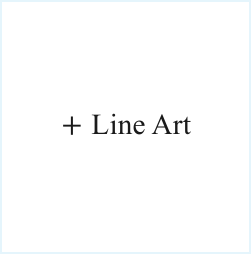}
        \vspace{-4mm}
        \caption*{Cond}
    \end{subfigure}\hfill
    \begin{subfigure}[t]{0.13\textwidth}
        \centering
        \includegraphics[width=\linewidth]{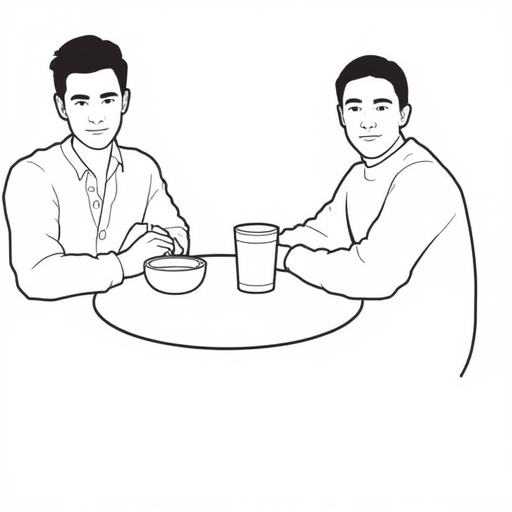}
        \vspace{-4mm}
        \caption*{(a)}
    \end{subfigure}\hfill
    \begin{subfigure}[t]{0.13\textwidth}
        \centering
        \includegraphics[width=\linewidth]{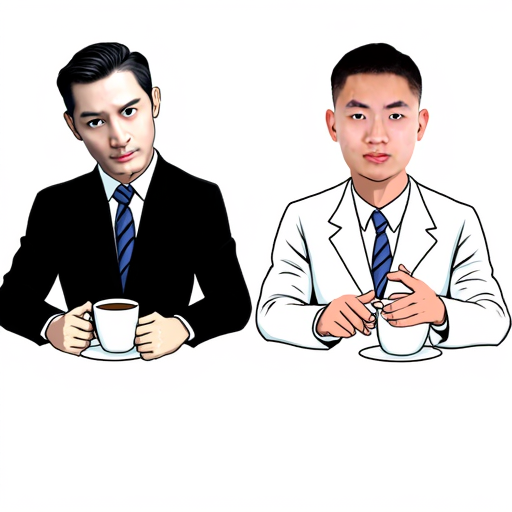}
        \vspace{-4mm}
        \caption*{(b)}
    \end{subfigure}\hfill
    \begin{subfigure}[t]{0.13\textwidth}
        \centering
        \includegraphics[width=\linewidth]{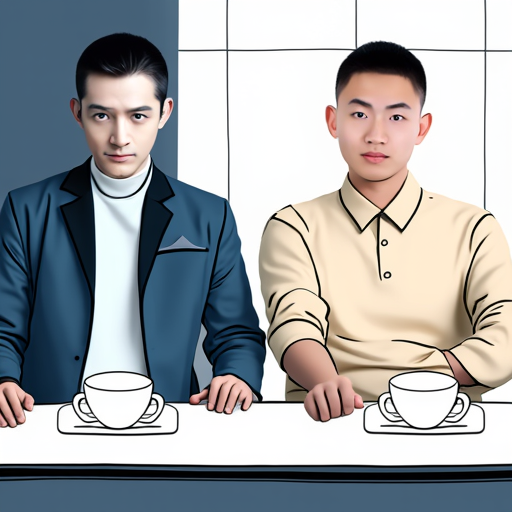}
        \vspace{-4mm}
        \caption*{(c)}
    \end{subfigure}\hfill
    \begin{subfigure}[t]{0.13\textwidth}
        \centering
        \includegraphics[width=\linewidth]{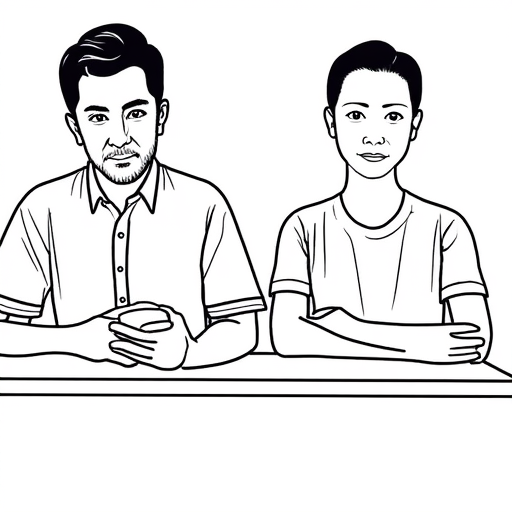}
        \vspace{-4mm}
        \caption*{(d)}
    \end{subfigure}\hfill
    \begin{subfigure}[t]{0.13\textwidth}
        \centering
        \includegraphics[width=\linewidth]{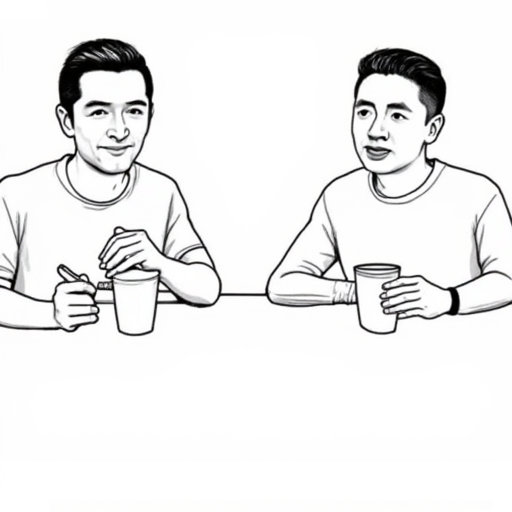}
        \vspace{-4mm}
        \caption*{\ours}
    \end{subfigure}\hfill
    \begin{subfigure}[t]{0.13\textwidth}
        \centering
        \includegraphics[width=\linewidth]{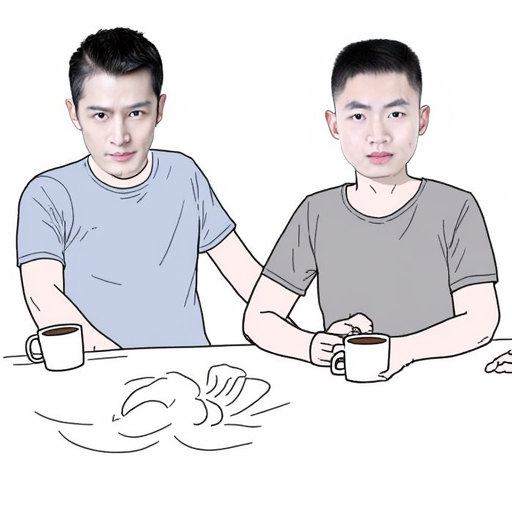}
        \vspace{-4mm}
        \caption*{(e)}
    \end{subfigure}

    \vspace{-2mm}
    \caption{Qualitative Ablations. 
    The first row shows generations without a specified style, while the second row demonstrates outputs conditioned on a ``Line Art'' style.}
    \label{fig:ablation_qual}
    \vspace{-8mm}
\end{figure*}

\paragraph{Setup.} \cref{tab:ablation} presents an incremental ablation on the MultiID-Bench 3--4-person subset, where we progressively add components of \ours. Variant (a) is the most basic baseline that directly uses face images as conditions, together with a ROI-mask for attention. In (b), we replace this design with our location canvas representation and apply location-aligned token pruning, which better anchors each identity to its target region while discarding irrelevant canvas tokens. Building on (b), (c) further introduces identity-adaptive modulation to inject identity features more effectively. Variant (d) additionally adopts our training strategies, including reference-face replacement and location image degradations, improving robustness to appearance gaps and imperfect spatial cues. Finally, (e) removes the modulation module from the full method, highlighting its role in preserving identity under multi-person conditioning.
\begin{wraptable}{r}{0.6\textwidth}
    \vspace{-11mm}
    \centering
        \caption{Ablation study of \ours on MultiID-Bench 3-and-4-person subset.}
        \scalebox{0.54}{
        \renewcommand{\arraystretch}{0.7}
        \begin{tabular}{lccccccc}
        \toprule
        \multirow{2}[1]{*}{\textbf{Variants}} & \multirow{2}[1]{*}{\textbf{Note}} & \multicolumn{3}{c}{\textbf{Identity Metrics}} & \multicolumn{3}{c}{\textbf{Generation Quality}} \\
        \cmidrule(r){3-5} \cmidrule(l){6-8}
         & & Sim(GT) $\uparrow$ & Sim(Ref) $\uparrow$ &  CP $\downarrow$ & CLIP-I $\uparrow$ & CLIP-T $\uparrow$ & Aes $\uparrow$ \\
        \midrule
        Base & (a) & 0.160 & 0.161 & \ignore 0.002 & 0.620 & 0.311 & 5.178 \\
        $\quad +$ location $\&$ pruning & (b) & 0.486 & 0.821 & 0.483 & 0.756 & 0.304 & 4.604 \\
        $\quad +$ modulation & (c) & 0.494 & 0.836 & 0.518 & 0.747 & 0.297 & 4.957 \\
        $\quad +$ replacement $\&$ aug. & (d) & 0.215 & 0.278 & \ignore 0.071 & 0.734 & 0.323 & 5.387 \\
        \rowcolor{ours} $\quad +$ $\mathcal{L}_{fs}$ (\ours) & -  & {0.424} & {0.587} & {0.186} & {0.763} & {0.326} & {4.960} \\
        $\quad -$ modulation & (e) & 0.365 & 0.583 & 0.276 & 0.741 & 0.317 & 5.260 \\
        \bottomrule
    \end{tabular}
        }       
    \label{tab:ablation}
    \vspace{-10mm}
\end{wraptable}%
Meanwhile, the visual impact of each component is further illustrated by the qualitative comparisons in~\cref{fig:ablation_qual}.

\paragraph{Analysis.}
We first observe that variant (a) yields low similarity to both GT and reference, and it also exhibits inaccurate spatial control and fragmented compositions (e.g., \cref{fig:ablation_qual}a-top). This corroborates our earlier discussion on the limitations of ROI-mask attention, which provides weak region grounding and tends to over-suppress useful context tokens. After introducing location-aligned token pruning in (b), both Sim(GT) and Sim(Ref) improve dramatically, and \cref{fig:ablation_qual}b-top shows precise location control. However, the CP score deteriorates sharply, leading to severe copy-paste behavior: when switching to a line-art prompt, the face region in \cref{fig:ablation_qual}b-bottom is largely insensitive to the target style and appears as a pasted face patch.
Adding identity-adaptive modulation in (c) further amplifies this issue: similarity scores increase, but CP becomes even higher, and \cref{fig:ablation_qual}c shows more obvious pasted-face artifacts. We attribute this to the fact that (c) is trained only with the flow-matching reconstruction loss $\mathcal{L}_{fm}$, which encourages pixel-level reconstruction; when the face condition is nearly identical to the target face, the modulation module is incentivized to emphasize reconstructing the entire input face, thus strengthening the copy-paste shortcut. To break this shortcut, (d) introduces reference-face replacement and location-canvas degradations. While these strategies effectively remove the one-to-one correspondence and keep the spatial control (\cref{fig:ablation_qual}d), the similarity scores are heavily affected, since $\mathcal{L}_{fm}$ alone provides no explicit supervision to relate the same person under different views/illumination/degradations, making the learning problem substantially harder.
Based on these findings, we incorporate the face similarity loss $\mathcal{L}_{fs}$ to form the full method. As shown in \cref{tab:ablation}, similarity recovers while CP is also suppressed, and \cref{fig:ablation_qual}\ours demonstrates both faithful identity preservation and prompt-driven style changes on the face region, effectively eliminating the copy-paste artifacts. Finally, removing identity-adaptive modulation from the full method (e) leads to lower similarity and higher CP, and \cref{fig:ablation_qual}e again exhibits copy-paste tendency. This indicates that $\mathcal{L}_{fs}$ alone is insufficient; the key is that identity-adaptive modulation can receive the $\mathcal{L}_{fs}$ feedback and selectively extract/amplify identity-relevant signals from conditions, which ultimately resolves copy-paste while maintaining identity.

\subsection{Further Analysis}
\label{sec:further_analysis}
\begin{figure*}[t]
    \centering
    \begin{subfigure}[t]{0.12\textwidth}
        \centering
        \includegraphics[width=\linewidth]{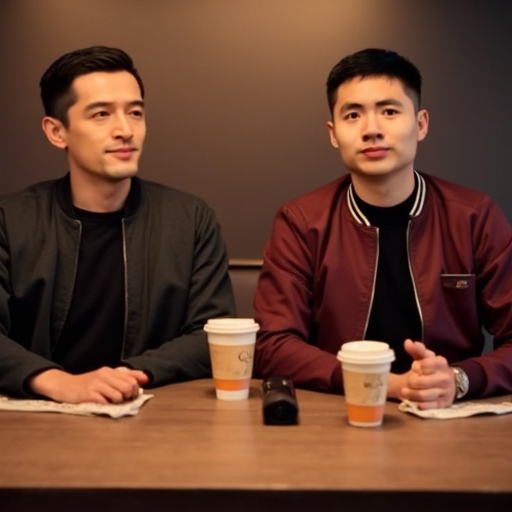}
        \vspace{-3mm}
        \caption*{Photoreal}
    \end{subfigure}\hfill
    \begin{subfigure}[t]{0.12\textwidth}
        \centering
        \includegraphics[width=\linewidth]{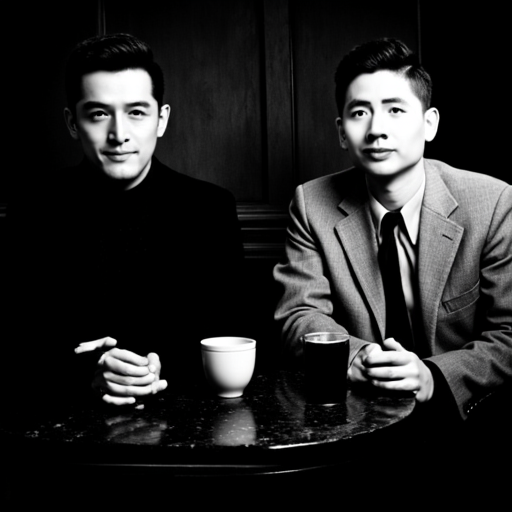}
        \vspace{-3mm}
        \caption*{Film noir}
    \end{subfigure}\hfill
    \begin{subfigure}[t]{0.12\textwidth}
        \centering
        \includegraphics[width=\linewidth]{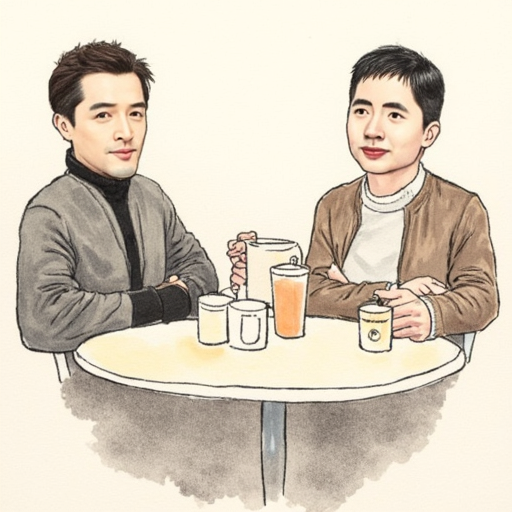}
        \vspace{-3mm}
        \caption*{Watercolor}
    \end{subfigure}\hfill
    \begin{subfigure}[t]{0.12\textwidth}
        \centering
        \includegraphics[width=\linewidth]{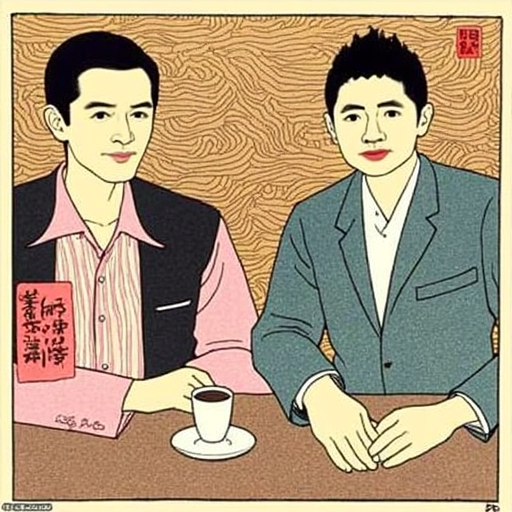}
        \vspace{-3mm}
        \caption*{Ukiyo-e}
    \end{subfigure}\hfill
    \begin{subfigure}[t]{0.12\textwidth}
        \centering
        \includegraphics[width=\linewidth]{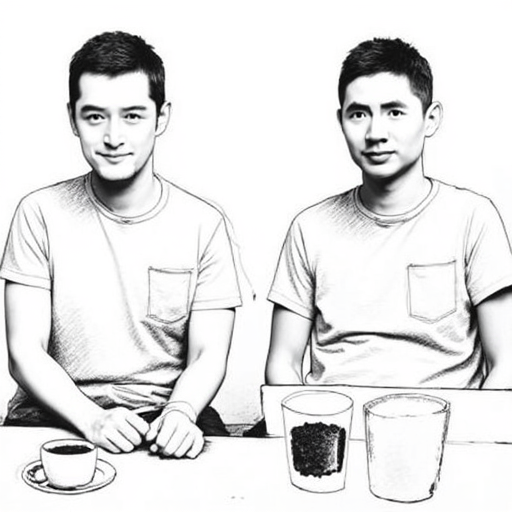}
        \vspace{-3mm}
        \caption*{Ink sketch}
    \end{subfigure}\hfill
    \begin{subfigure}[t]{0.12\textwidth}
        \centering
        \includegraphics[width=\linewidth]{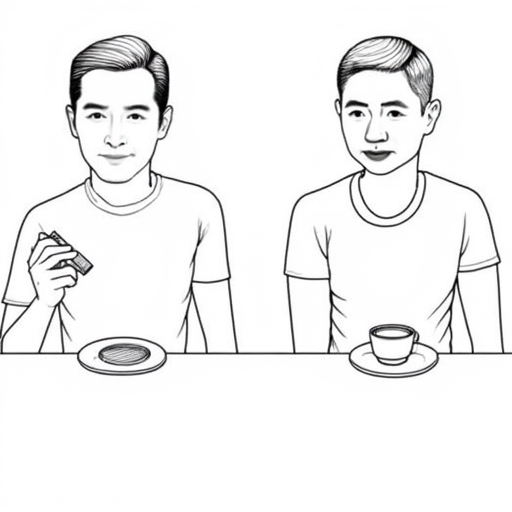}
        \vspace{-3mm}
        \caption*{Line art}
    \end{subfigure}\hfill
    \begin{subfigure}[t]{0.12\textwidth}
        \centering
        \includegraphics[width=\linewidth]{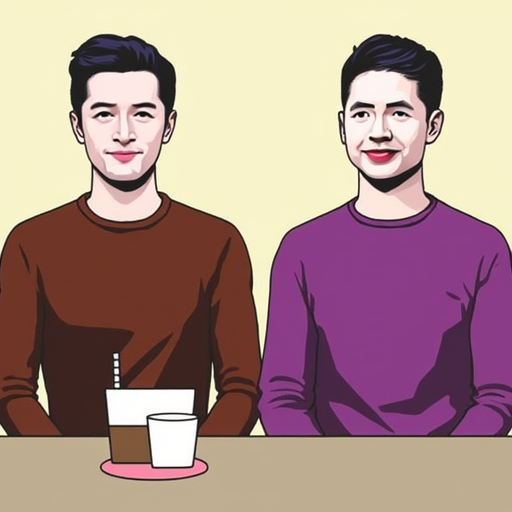}
        \vspace{-3mm}
        \caption*{Flat vector}
    \end{subfigure}\hfill
    \begin{subfigure}[t]{0.12\textwidth}
        \centering
        \includegraphics[width=\linewidth]{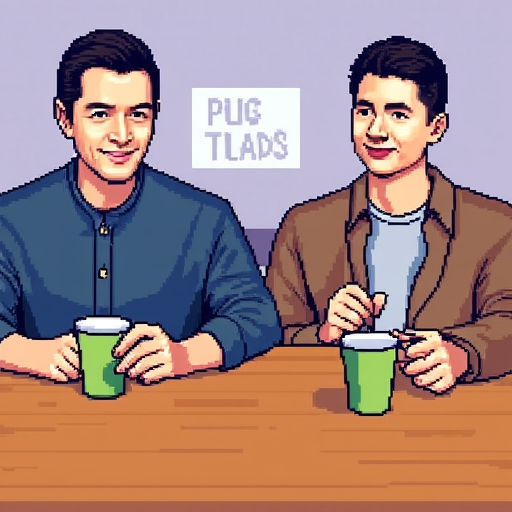}
        \vspace{-3mm}
        \caption*{Pixel art}
    \end{subfigure}

    \vspace{-2mm}
    \caption{\ours is capable of generating high-quality images across diverse styles while preserving identity and enforcing accurate location control.}
    \label{fig:style_results}
    \vspace{-7mm}
\end{figure*}

\paragraph{Style Diversity.}
Beyond photorealistic rendering, \ours supports prompt-driven stylization while maintaining both identity consistency and spatial controllability. \cref{fig:style_results} shows that the same set of identities can be faithfully preserved under a wide range of artistic styles (e.g., watercolor, ukiyo-e, ink sketch, and pixel art). Importantly, the stylization effect is applied coherently to the face region instead of being overridden by identity conditions, indicating that our design avoids the common copy-paste shortcut and enables compositional generation where identity cues and text styles jointly shape the final appearance.

\begin{wrapfigure}{r}{0.4\textwidth}
    \vspace{-7mm}
    \centering
    \begin{subfigure}[t]{0.13\textwidth}
        \centering
        \includegraphics[width=\linewidth]{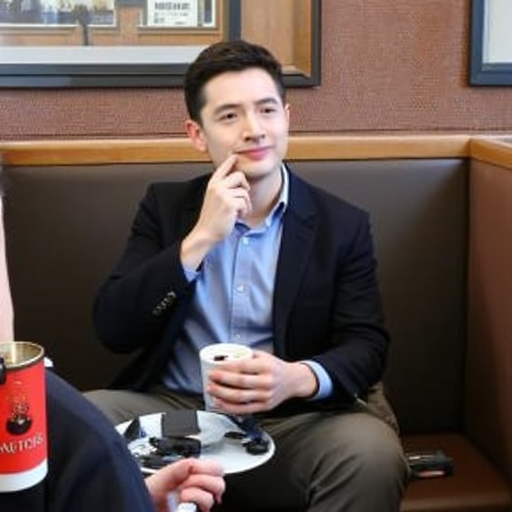}
        \vspace{-3mm}
        \caption*{Overlap}
    \end{subfigure}\hfill
    \begin{subfigure}[t]{0.13\textwidth}
        \centering
        \includegraphics[width=\linewidth]{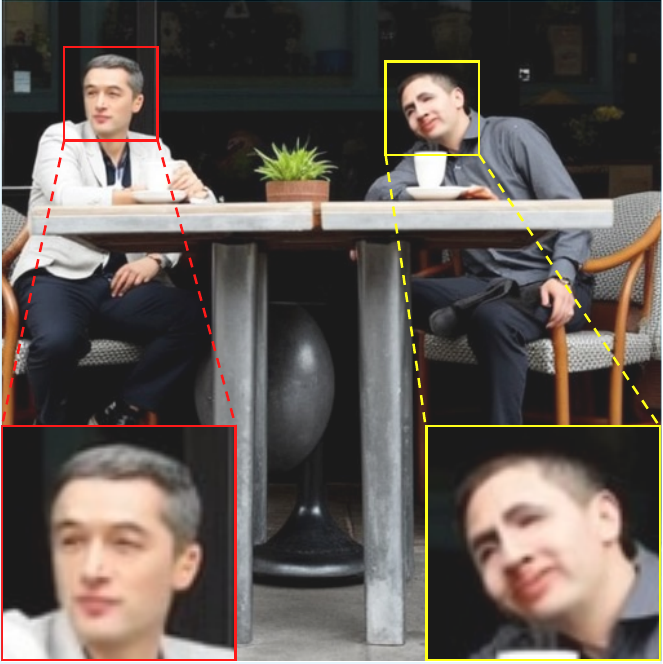}
        \vspace{-3mm}
        \caption*{Small face}
    \end{subfigure}\hfill
    \begin{subfigure}[t]{0.13\textwidth}
        \centering
        \includegraphics[width=\linewidth]{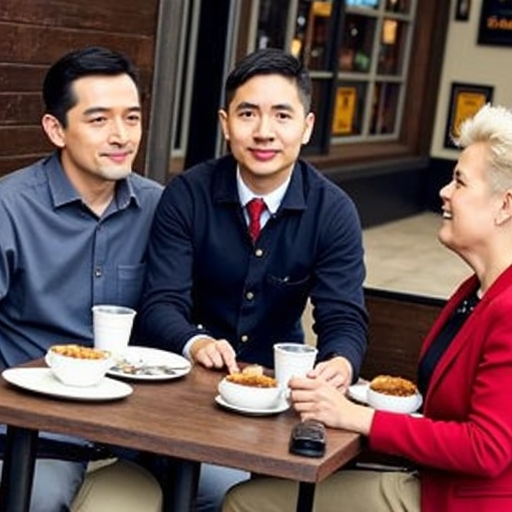}
        \vspace{-3mm}
        \caption*{Conflict}
    \end{subfigure}\hfill

    \vspace{-2mm}
    \caption{Failure cases example of extreme conditioning.}
    \label{fig:failure_cases}
    \vspace{-7mm}
\end{wrapfigure}

\paragraph{Limitations/Failure Cases.}
Despite strong overall performance, \ours can fail under extreme conditions (\cref{fig:failure_cases}). When target regions heavily overlap,
different identities may compete for the same spatial tokens, leading to identity mixing or local swaps. When the input face is extremely small, fine-grained facial details (e.g., eyes) become blurry and the synthesized head may be less compatible with the generated body due to insufficient identity evidence and weak spatial supervision at the effective token resolution. Moreover, conflicting or under-specified prompts can make the generation highly unstable---for example, providing two reference faces but asking the model to generate three people often yields uncontrollable identity assignment. Addressing these corner cases likely requires more research in future work.

\section{Related Works}
\label{sec:related}
\paragraph{Identity-Preserving Image Generation}
aims to synthesize images that follow a text prompt while faithfully retaining the identity characteristics of a given reference subject. Early studies mainly focus on 
\emph{single-subject} personalization~\cite{gal2022image,valevski2023face0,hu2022lora,ruiz2023dreambooth,wang2024instantid,ye2023ip,li2024photomaker,tao2025instantcharacter,jiang2025infiniteyou}. 
A representative line is \emph{test-time tuning}, where model parameters or embeddings are optimized per identity. For example, Textual Inversion~\cite{gal2022image} learns a dedicated identity token via prompt/embedding optimization, and its idea of injecting identity information in the text/embedding space inspires later works such as Face0~\cite{valevski2023face0}, PhotoMaker~\cite{li2024photomaker}, and FaceStudio~\cite{yan2023facestudio}. 
Despite strong fidelity, test-time tuning is often inflexible in practice since it requires per-identity optimization and is costly when the number of identities grows.
To improve flexibility, \emph{training-time tuning} and \emph{tuning-free} methods learn a general mapping from reference images to identity features, which can be plugged into the generator at inference. 
Representatively, IP-Adapter~\cite{ye2023ip} encodes reference images into identity features via an image encoder and an MLP, and injects them through cross-attention. 
InstantID~\cite{wang2024instantid} further incorporates a dedicated face recognition model for identity extraction and introduces ControlNet-like control to focus identity injection on the face region. 
InstantCharacter~\cite{tao2025instantcharacter} emphasizes holistic subject preservation by combining global and local reference features, while InfiniteYou~\cite{jiang2025infiniteyou} integrates face keypoints and identity features through an InfuseNet to enhance robustness.
Recently, identity-preserving generation has been extended to the significantly more challenging \emph{multi-subject} setting~\cite{zhou2024storymaker,xiao2025fastcomposer,kim2024instantfamily,he2025uniportrait,he2025anystory,mou2025dreamo,zhang2025id,xu2025withanyone,qian2025layercomposer,borse2026ar2canarchitectartistleveraging}, where multiple reference identities must be preserved simultaneously.
One major direction follows \emph{identity feature extraction plus ROI-aware injection}. StoryMaker~\cite{zhou2024storymaker} and WithAnyOne~\cite{xu2025withanyone} extract identity features from references and inject them via ROI-attention to control where each identity appears. 
FastComposer~\cite{xiao2025fastcomposer} and InstantFamily~\cite{kim2024instantfamily} fuse identity features into the text stream and combine them with ROI-attention for localized control. 
UniPortrait~\cite{he2025uniportrait} and AnyStory~\cite{he2025anystory} further replace fixed ROI masks with learnable routing modules, allowing the model to automatically determine the spatial assignment of identities.
Another line is inspired by OminiControl~\cite{tan2025ominicontrol} and leverages \emph{aligned RoPE} to anchor identity conditions to specific spatial tokens~\cite{xu2025contextgen,qian2025layercomposer}. 
ContextGen~\cite{xu2025contextgen} proposes a contextual canvas (context image) where conditions are pasted onto an image-sized layout, and RoPE alignment is used to associate conditions with target regions. 
LayerComposer~\cite{qian2025layercomposer} builds upon this idea and introduces transparent-token cropping to better utilize spatial tokens. 
More recently, UMO~\cite{cheng2025umo} and Ar2Can~\cite{borse2026ar2canarchitectartistleveraging} explore incorporating reinforcement learning objectives into multi-human generation.
Our method is heavily inspired by LayerComposer~\cite{qian2025layercomposer} and XVerse~\cite{chen2025xverse}, while differing in key designs. 
Compared to LayerComposer, we focus on the most identity-critical \emph{face} region of the reference (instead of full-body cues) and treat all reference faces symmetrically when designing RoPE alignment, rather than using asymmetric handling for different references. 
Additionally, we modulate \emph{condition tokens} using \emph{face features} so that the model can adaptively select, attend to, and amplify identity-relevant signals from the conditions, whereas XVerse modulates the \emph{text token} corresponding to the identity to strengthen textual control. These differences reflect distinct design goals and lead to substantially different mechanisms for multi-identity control.

\section{Conclusion}
We study identity-preserving image generation with explicit location control, where existing approaches often suffer from inaccurate spatial grounding or shortcut behaviors such as copy--paste. To address these issues, we propose \ours, a simple yet effective framework built on diffusion transformers.
Our key insight is to disentangle identity binding from free-form generation: we introduce identity-isolated attention to prevent identity leakage across regions, adopt location-aligned token pruning to strengthen spatial controllability, and design an id-adaptive modulation mechanism trained with a face-similarity objective to suppress copy--paste while maintaining high identity fidelity.
Extensive experiments demonstrate that \ours consistently improves both quantitative metrics and visual quality, supports diverse styles under text prompts while preserving identity, and enables accurate position-controlled synthesis. We hope this work offers a practical step toward reliable multi-identity generation.

%
%
\bibliographystyle{splncs04}
\bibliography{main}

\clearpage

\appendix

\renewcommand{\thetable}{\Alph{table}} 
\setcounter{table}{0}                 

\renewcommand{\thefigure}{\Alph{figure}}
\setcounter{figure}{0}

\section{Implementation Details}
To facilitate reproducibility, we provide additional implementation details and hyperparameters that complement the main text.

\subsection{Learnable Modules}
We build \ours upon a frozen FLUX.1-[dev] backbone and only introduce lightweight learnable components: (i) LoRA adapters attached to the image branch to handle the newly introduced location-reference tokens, and (ii) a small MLP that projects a face-recognition embedding into an identity modulation offset for AdaLN-style conditioning.

\paragraph{LoRA.} We use FLUX.1-[dev]~\cite{flux2024} as the base model. To accommodate the newly introduced location-reference tokens (i.e., VAE tokens encoded from the location canvas), we attach LoRA adapters (rank $r{=}128$) to the linear layers along the same forward path used by the noised latent tokens. The target modules are selected by the following regex-based matching rules:
\begin{lstlisting}[language=Python]
regex_pattern = (
    r".*x_embedder|"
    r".*(?<!single_)transformer_blocks\.[0-9]+\.norm1\.linear|"
    r".*(?<!single_)transformer_blocks\.[0-9]+\.attn\.to_k|"
    r".*(?<!single_)transformer_blocks\.[0-9]+\.attn\.to_q|"
    r".*(?<!single_)transformer_blocks\.[0-9]+\.attn\.to_v|"
    r".*(?<!single_)transformer_blocks\.[0-9]+\.attn\.to_out\.0|"
    r".*(?<!single_)transformer_blocks\.[0-9]+\.ff\.net\.2|"
    r".*single_transformer_blocks\.[0-9]+\.norm\.linear|"
    r".*single_transformer_blocks\.[0-9]+\.proj_mlp|"
    r".*single_transformer_blocks\.[0-9]+\.proj_out|"
    r".*single_transformer_blocks\.[0-9]+\.attn\.to_k|"
    r".*single_transformer_blocks\.[0-9]+\.attn\.to_q|"
    r".*single_transformer_blocks\.[0-9]+\.attn\.to_v|"
    r".*single_transformer_blocks\.[0-9]+\.attn\.to_out"
)
\end{lstlisting}
To avoid degrading the original generation capability, LoRA is enabled \emph{only} when processing the location-reference tokens; it is disabled for the noised latent tokens.

\paragraph{Face Embedding Projector.} As described in Section 3.3, we extract an identity embedding $e_{r_i}$ for each reference face $r_i$ using a pretrained face recognition model (e.g., ArcFace~\cite{deng2019arcface}). We map $e_{r_i}$ to an additive modulation offset $\Delta_{F_i}$ via a lightweight MLP and inject identity through the modulation pathway by using $y_i = y + \Delta_{F_i}$ \emph{only for tokens originating from the corresponding identity branch} $F_i$, where $y$ is the original global conditioning vector from timestep and text.

Concretely, the projector is a 2-layer MLP with the structure \emph{Linear} $\rightarrow$ \emph{GELU} $\rightarrow$ \emph{Linear}, whose output dimension matches the DiT conditioning dimension (the same dimensionality as $y$). This design keeps the identity pathway lightweight while providing a persistent per-block identity bias through AdaLN-style modulation.

\subsection{Data Processing}

\paragraph{Pre-processing.} Our data preprocessing pipeline contains four stages. (1) \textbf{Instance segmentation for human detection.} Given an input image, we first detect each person instance using a YOLO11x~\cite{yolo11_ultralytics} model, producing person masks (or bounding boxes) to localize individual subjects. (2) \textbf{Face detection and facial geometry.} For each detected person instance, we run face detection using \texttt{insightface} (\emph{antelopev2} method)~\cite{Deng2020CVPR}. We extract (i) the face bounding box and (ii) five facial landmarks following the ArcFace 5-point protocol~\cite{deng2019arcface}. We use the bounding box in the following matching step and the landmarks as the face location $l_i$ for training-time processing. (3) \textbf{Reference matching.} To assign each detected face to a known identity from MultiID-2M~\cite{xu2025withanyone}, we compute a weighted matching score that combines face-box IoU and face-embedding similarity, and perform greedy matching to obtain the reference identity mapping (\texttt{id-name}) for each face. (4) \textbf{Image annotation.} Finally, we use Qwen-2.5-VL-72B~\cite{bai2025qwen25vltechnicalreport} to generate textual annotations for each image.
The prompt for Qwen-2.5-VL-72B is as follows:
\begin{lstlisting}[language=Python]
"""
Generate a vivid, natural scene description as if you're directly observing it. 
Avoid phrases like "this picture" or "in this image" or "this scene" or "the content of". Instead, immerse the reader in the scene by:
- Focusing on subjects and their attributes (clothing, appearance, actions and so on)
- Capturing meaningful spatial relationships
- Maintaining a flowing narrative style
- Environment and setting details
- Use present continuous tense for actions
- Be descriptive but concise
Example one: "A young couple is embracing near a waterfront, with the woman resting her head on the man's shoulder as they watch the sunset."
Example two: "A group of friends is picnicking on a grassy lawn, with two women laughing as a man pours drinks from a thermos, while children play with a frisbee in the background."
Example three: "A young woman in a floral dress is smiling while holding a bouquet of flowers in a sunlit garden."
The goal is to describe vividly and directly with descriptive and concise language.
"""
\end{lstlisting}

\paragraph{Training Data Pipeline.}
During training, we apply the following pipeline to construct paired supervision for identity injection and location conditioning. (1) \textbf{Norm-cropped face extraction.} For each detected face location $l_i$ (ArcFace 5-point landmarks), we extract a normalized face crop $r_i$ (\emph{norm-crop}). (2) \textbf{Reference face replacement.} Based on the matched \texttt{id-name} from the pre-processing stage, we load the corresponding reference face $\tilde{r}_i$ from the reference set and, with probability $p$, replace $r_i$ by $\tilde{r}_i$. We adopt a curriculum schedule where $p$ is increased from 0 to 0.5 following $[0.0, 0.05, 0.1, 0.2, 0.3, 0.4, 0.5]$ at milestone steps $[10\mathrm{k}, 20\mathrm{k}, 30\mathrm{k}, 40\mathrm{k}, 50\mathrm{k}, 60\mathrm{k}]$. (3) \textbf{Square-preserving image crop.} We apply a square crop to fit the training resolution, while ensuring that \emph{all} faces of interest remain inside the cropped region. (4) \textbf{Location canvas construction with face degradations.} We construct the location canvas used to produce location-reference tokens, and randomly apply degradations to each face region on the canvas to simulate imperfect references, including ColorJitter, Grayscale, Blur, Sharpness, Autocontrast, horizontal flip, salt-and-pepper noise, Gaussian noise, occlusion, illumination changes, JPEG artifacts, resolution degradation, and face aging.

\subsection{Location Canvas Construction}
\label{sec:appendix_location_canvas}

We construct a \emph{location canvas} by pasting a (possibly degraded) norm-cropped face patch onto a blank image-sized canvas at a location determined by facial landmarks. The resulting canvas is then encoded by the VAE to obtain the location-reference tokens.

\paragraph{Notation.} Let $I\in\mathbb{R}^{H\times W\times 3}$ be the (square-cropped) training image and $L\in\mathbb{R}^{H\times W\times 3}$ be an initially blank \emph{white} canvas (all pixels set to 255 in RGB space). For each face $i$, we have a face patch $f_i$ (canonical size $112\times112$), its 5-point landmarks in the patch coordinate system $K_i^{\text{src}}\in\mathbb{R}^{5\times 2}$, and the corresponding 5-point landmarks in the image/crop coordinate system $K_i^{\text{dst}}\in\mathbb{R}^{5\times 2}$ (ArcFace protocol).
We also maintain an alpha map $a_i\in[0,255]^{112\times112}$; if not provided, we use a fully opaque alpha.

\paragraph{Scale estimation (SimilarityTransform).} We estimate a similarity transform between landmarks using \texttt{SimilarityTransform}~\cite{van_der_Walt_2014}. Let the estimated affine parameters be
\begin{equation}
T = \begin{bmatrix} \mathbf{A} & \mathbf{t} \\\ 0 & 1 \end{bmatrix},\quad \mathbf{A}\in\mathbb{R}^{2\times 2},\ \mathbf{t}\in\mathbb{R}^{2},
\end{equation}
where for a true similarity transform, $\mathbf{A}=s\mathbf{R}$ with scale $s$ and rotation $\mathbf{R}$. In implementation, the scale is computed from the linear part as
\begin{equation}
\hat{s} = \sqrt{\left|\det(\mathbf{A})\right|}.
\end{equation}
Due to the argument ordering of the estimation call, we use the reciprocal to obtain the patch-to-image scale and clamp it to a reasonable range:
\begin{equation}
 s_i = \mathrm{clip}\left(\frac{1}{\hat{s}},\ 0.2,\ 5.0\right).
\end{equation}
We then resize the face patch and alpha map to $(112\,s_i)\times(112\,s_i)$ (alpha uses nearest-neighbor resizing).

\paragraph{Translation by mean landmark residual.} After scaling the source landmarks by $s_i$, we compute a translation offset as the mean residual over the 5 landmarks:
\begin{equation}
\Delta_i \,=\, \frac{1}{5}\sum_{j=1}^{5}\left(K_{i,j}^{\text{dst}} - s_i K_{i,j}^{\text{src}}\right) \in \mathbb{R}^{2}.
\end{equation}
In practice we use integer offsets $(\lfloor\Delta_{i,x}\rfloor,\lfloor\Delta_{i,y}\rfloor)$ as the paste position on the image-sized canvas.

\paragraph{Alpha compositing and binary mask.} We maintain a global mask $M\in\{0,1\}^{H\times W}$ initialized as zeros. For each face, we paste the resized alpha map into $M$ at the computed offset and binarize it:
\begin{equation}
M(x,y) \leftarrow \mathbb{I} \big[M(x,y) > 0\big],
\end{equation}
which matches the implementation that thresholds the pasted alpha to obtain a binary location mask. Finally, we paste the resized face patch onto the canvas using alpha blending (PIL alpha compositing), i.e., pixels covered by higher alpha overwrite more strongly. The output of this stage is the location canvas $L$ and its mask $M$.

\paragraph{Remarks.} If the pasted region exceeds image boundaries, the paste operation is automatically clipped by the image library. The face-region degradations described in the training pipeline are applied on $f_i$ (and/or $a_i$) before pasting, so that the location reference remains robust to imperfect observations.

\subsection{Metrics}
\label{sec:appendix_metrics}

\paragraph{Overview.}
We evaluate on MultiID-Bench~\cite{xu2025withanyone}, a unified benchmark for group-photo (multi-ID) generation with rare long-tail identities that do not overlap with training data. Each test case contains a ground-truth (GT) image with 1--4 people, the corresponding 1--4 reference images as inputs, and a prompt describing the GT. We report both \emph{identity fidelity} and \emph{generation quality}.

\paragraph{Face embedding and cosine similarity.}
For each identity $i$ in a test case, we compute face embeddings for the reference, GT, and generated faces, denoted as $\mathbf r_i,\mathbf t_i,\mathbf g_i$, respectively.
We define the similarity between two embeddings by cosine similarity
\begin{equation}
\mathrm{Sim}(\mathbf a,\mathbf b) = \frac{\mathbf a^\top \mathbf b}{\|\mathbf a\|\,\|\mathbf b\|}.
\end{equation}
We then define
\begin{equation}
\mathrm{Sim(Ref)} = \mathrm{Sim}(\mathbf r_i,\mathbf g_i),\qquad
\mathrm{Sim(GT)}  = \mathrm{Sim}(\mathbf t_i,\mathbf g_i).
\end{equation}
While many prior works mainly report $\mathrm{Sim_{Ref}}$, it can inadvertently favor trivial copy-paste---directly replicating the reference maximizes the score even when the prompt implies changes in pose, expression, or viewpoint. Therefore, we use $\mathrm{Sim_{GT}}$ as the primary identity metric and additionally report $\mathrm{Sim_{Ref}}$ to quantify reference consistency.

\paragraph{Copy-paste tendency (CP).}
To explicitly measure whether a method collapses to directly pasting the reference appearance, we compute the angular distance on the unit sphere,
\begin{equation}
\theta_{ab} = \arccos\big(\mathrm{Sim}(\mathbf a,\mathbf b)\big).
\end{equation}
The copy-paste metric is defined as
\begin{equation}
\mathrm{CP}(\mathbf g_i\mid \mathbf t_i,\mathbf r_i)
=\frac{\theta_{g_it_i}-\theta_{g_ir_i}}{\max(\theta_{t_ir_i},\,\varepsilon)},
\end{equation}
where $\varepsilon$ is a small constant for numerical stability.
A larger CP indicates that the generated identity is biased toward the reference (copy-paste), while a smaller (more negative) CP indicates closer agreement with the GT identity.

\paragraph{Prompt fidelity and aesthetics.}
We further report generation quality using standard metrics: CLIP text alignment (CLIP-T) between the generated image and the prompt, CLIP image alignment (CLIP-I) between the generated image and the GT image, and an off-the-shelf aesthetic score (Aes).

\begin{table}[h]
    \centering
        \caption{Effect of Identity-Isolated Attention.}
        \vspace{-2mm}
        \scalebox{0.8}{
        \renewcommand{\arraystretch}{0.8}
        \begin{tabular}{lcccccc}
        \toprule
        \multirow{2}[1]{*}{\textbf{Variants}} & \multicolumn{3}{c}{\textbf{Identity Metrics}} & \multicolumn{3}{c}{\textbf{Generation Quality}} \\
        \cmidrule(r){2-4} \cmidrule(l){5-7}
         & Sim(GT) $\uparrow$ & Sim(Ref) $\uparrow$ &  CP $\downarrow$ & CLIP-I $\uparrow$ & CLIP-T $\uparrow$ & Aes $\uparrow$ \\
        \midrule
        All visible & 0.353 & 0.414 & 0.113 & 0.746 & 0.302 & 4.764 \\
        Identity-Isolated (\ours) & {0.424} & {0.587} & {0.186} & {0.763} & {0.326} & {4.960} \\
        \bottomrule
    \end{tabular}
        }       
    \label{tab:effect_of_identity_isolated_attention}
\end{table}%

\section{Further Experiments}
\paragraph{Effect of Identity-Isolated Attention.}
We ablate the proposed star-topology attention mask by allowing all identity branches to attend to each other (\emph{All visible}), which can introduce cross-identity interference when multiple reference faces co-exist in the same sequence. As shown in Table~\ref{tab:effect_of_identity_isolated_attention}, Identity-Isolated Attention substantially improves identity fidelity and overall quality: Sim(GT) increases from 0.353 to 0.424 (+0.071) and Sim(Ref) from 0.414 to 0.587 (+0.173), accompanied by higher CLIP-I (0.746$\rightarrow$0.763), CLIP-T (0.302$\rightarrow$0.326), and aesthetics (4.764$\rightarrow$4.960). Although CP becomes larger (0.113$\rightarrow$0.186), indicating a stronger pull toward the reference, the simultaneous gain in Sim(GT) suggests that isolating identity branches primarily mitigates identity leakage/blending and stabilizes multi-person composition rather than merely encouraging trivial copy-paste.

\begin{table}[h]
    \centering
        \caption{Effect of curriculum reference face replacement.}
        \vspace{-2mm}
        \scalebox{0.8}{
        \renewcommand{\arraystretch}{0.8}
        \begin{tabular}{lcccccc}
        \toprule
        \multirow{2}[1]{*}{\textbf{Variants}} & \multicolumn{3}{c}{\textbf{Identity Metrics}} & \multicolumn{3}{c}{\textbf{Generation Quality}} \\
        \cmidrule(r){2-4} \cmidrule(l){5-7}
         & Sim(GT) $\uparrow$ & Sim(Ref) $\uparrow$ &  CP $\downarrow$ & CLIP-I $\uparrow$ & CLIP-T $\uparrow$ & Aes $\uparrow$ \\
        \midrule
        Fixed $p=0.5$ & 0.403 & 0.529 & 0.141 & 0.762 & 0.316 & 4.871 \\
        Curriculum (\ours) & {0.424} & {0.587} & {0.186} & {0.763} & {0.326} & {4.960} \\
        \bottomrule
    \end{tabular}
        }       
    \label{tab:effect_of_curriculum_reference_face_replacement}
\end{table}%

\paragraph{Effect of Curriculum Reference Face Replacement.}
We compare using a fixed reference-face replacement probability ($p{=}0.5$ throughout training) with our curriculum that gradually increases $p$ (see the training pipeline in Appendix~\ref{sec:appendix_location_canvas}). As shown in Table~\ref{tab:effect_of_curriculum_reference_face_replacement}, the curriculum yields consistently better identity fidelity and overall generation quality: it improves Sim(GT) from 0.403 to 0.424 (+0.021) and Sim(Ref) from 0.529 to 0.587 (+0.058), and also increases CLIP-T (0.316$\rightarrow$0.326) and aesthetics (4.871$\rightarrow$4.960). We observe a mild increase in CP (0.141$\rightarrow$0.186), suggesting a slightly stronger reference bias; nevertheless, the higher Sim(GT) indicates that the curriculum better balances leveraging reference identities while still matching the ground-truth identity implied by the prompt.

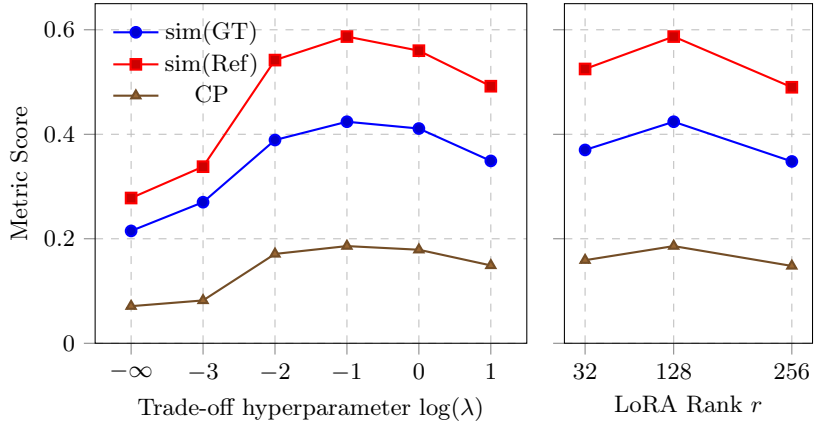
\begin{figure}[t]
\centering
\begin{tikzpicture}
\begin{groupplot}[
    group style={
        group size=2 by 1,
        horizontal sep=0.5cm,
        vertical sep=0cm,
    },
    height=0.5\textwidth, 
    ymin=0, ymax=0.65,
    grid=both,
    grid style={dashed},
    tick align=outside,
    tick pos=left,
    legend style={draw=none, fill=none},
]

\nextgroupplot[
    xlabel={Trade-off hyperparameter $\log(\lambda)$},
    ylabel={Metric Score},
    width=0.6\textwidth,
    xmode=log,
    log basis x=10,
    xtick={0.0001,0.001,0.01,0.1,1,10},
    xticklabels={$-\infty$,$-3$,$-2$,$-1$,$0$,$1$},
    legend style={at={(0.02,0.98)}, anchor=north west},
]

\addplot+[mark=*, thick] coordinates {
    (0.0001,0.215) (0.001,0.270) (0.01,0.389) (0.1,0.424) (1,0.411) (10,0.349)
};
\addlegendentry{sim(GT)}

\addplot+[mark=square*, thick] coordinates {
    (0.0001,0.278) (0.001,0.338) (0.01,0.542) (0.1,0.587) (1,0.560) (10,0.492)
};
\addlegendentry{sim(Ref)}

\addplot+[mark=triangle*, thick] coordinates {
    (0.0001,0.071) (0.001,0.082) (0.01,0.171) (0.1,0.186) (1,0.179) (10,0.149)
};
\addlegendentry{CP}

\nextgroupplot[
    width=0.4\textwidth,
    xlabel={LoRA Rank $r$},
    xtick={32,128,256},
    xticklabels={$32$,$128$,$256$},
    ytick={},
    yticklabels={}
]

\addplot+[mark=*, thick] coordinates {
    (32,0.370) (128,0.424) (256,0.348)
};

\addplot+[mark=square*, thick] coordinates {
    (32,0.525) (128,0.587) (256,0.490)
};

\addplot+[mark=triangle*, thick] coordinates {
    (32,0.159) (128,0.186) (256,0.148)
};

\end{groupplot}
\end{tikzpicture}

\caption{Sensitivity to the trade-off hyperparameter $\lambda$ and LoRA Rank $r$.}
\label{fig:lambda_sensitivity}
\end{figure}

\paragraph{Sensitivity to Trade-off hyperparameter $\lambda$.}
We analyze the sensitivity to the trade-off weight $\lambda$ in the training objective
$\mathcal{L}=\mathcal{L}_{fm}+\lambda\,\mathcal{L}_{fs}$ (Section~3.5).
As shown in Fig.~\ref{fig:lambda_sensitivity} (left), very small $\lambda$ provides insufficient
identity-related supervision, resulting in reduced identity fidelity.
Increasing $\lambda$ strengthens identity consistency and improves identity metrics, with the
best overall trade-off achieved at a moderate value.
Meanwhile, the copy-paste tendency generally increases with larger $\lambda$, but remains
tolerable (CP $<0.2$) and does not exhibit a pronounced copy-paste effect.
When $\lambda$ becomes too large, the model is over-regularized and performance deteriorates,
suggesting that overly emphasizing $\mathcal{L}_{fs}$ can hinder faithful generation.
Unless otherwise specified, we set $\lambda=0.1$ in all experiments.

\paragraph{Sensitivity to LoRA Rank $r$.}
As shown in Fig.~\ref{fig:lambda_sensitivity} (right), varying the LoRA rank $r$ yields a clear capacity--generalization trade-off. With a small rank ($r{=}32$), the adapter capacity is insufficient to model the newly introduced location-reference conditioning pathway, leading to underfitting and consistently weaker identity fidelity and overall generation quality. Increasing $r$ improves adaptation and stabilizes training, but overly large ranks ($r{=}256$) introduce excessive capacity relative to our training data scale, making the model prone to overfitting and thus hurting generalization at evaluation time. Overall, a moderate rank provides the best balance; unless otherwise specified, we use $r{=}128$ in all experiments.

\section{Qualitative Results}

\begin{figure}
    \centering
    \includegraphics[width=\textwidth]{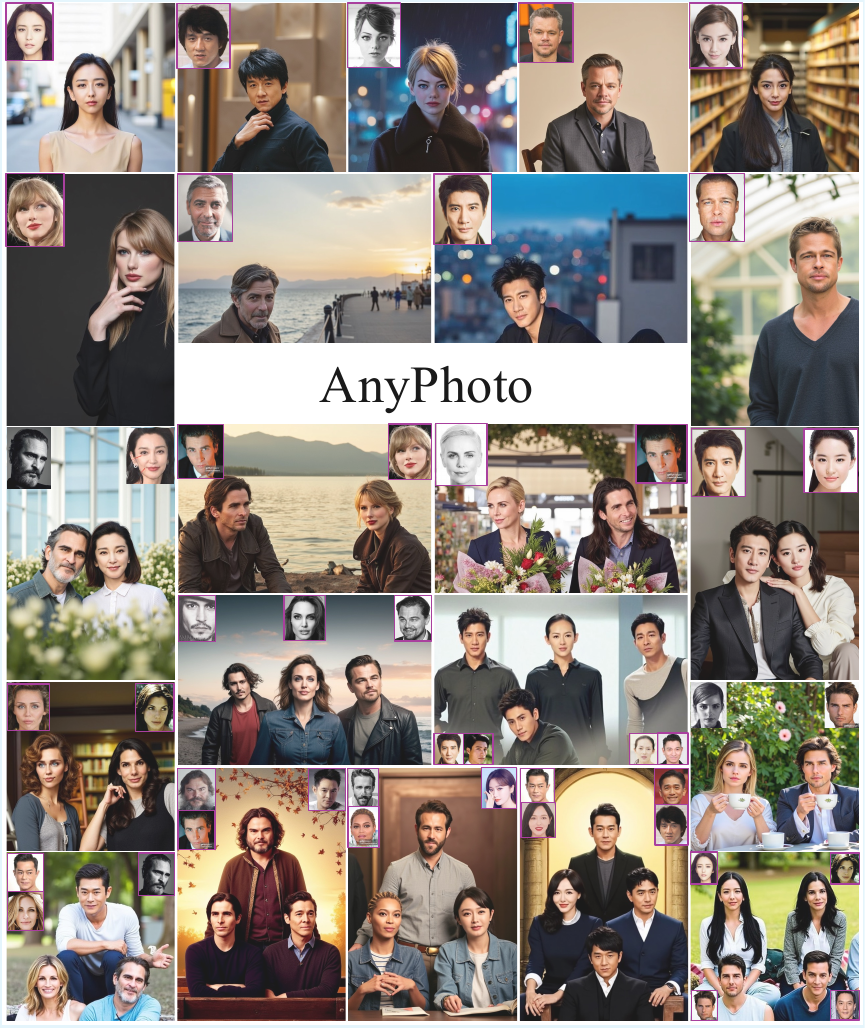}
    \caption{Additional qualitative results of \ours. Across diverse scenes and 1--4 identity conditions, \ours preserves each identity while enforcing accurate location control, producing coherent compositions with fewer copy-paste artifacts.}
\end{figure}

\end{document}